\documentclass[10pt,journal,compsoc]{IEEEtran}

\usepackage{xr}
\makeatletter
\newcommand*{\addFileDependency}[1]{
  \typeout{(#1)}
  \@addtofilelist{#1}
  \IfFileExists{#1}{}{\typeout{No file #1.}}
}
\makeatother

\newcommand*{\myexternaldocument}[1]{%
    \externaldocument{#1}%
    \addFileDependency{#1.tex}%
    \addFileDependency{#1.aux}%
}
\myexternaldocument{supplementary}

\usepackage{hyperref}
\hypersetup{
    unicode=false,          
    pdftoolbar=true,        
    pdfmenubar=true,        
    pdffitwindow=false,     
    pdfstartview={FitH},    
    pdftitle={My title},    
    pdfauthor={Author},     
    pdfsubject={Subject},   
    pdfcreator={Creator},   
    pdfproducer={Producer}, 
    pdfkeywords={keyword1, key2, key3}, 
    pdfnewwindow=true,      
    colorlinks=false,       
    linkcolor=red,          
    citecolor=green,        
    filecolor=magenta,      
    urlcolor=cyan,           
}

\usepackage{adjustbox}
\usepackage{xcolor}
\usepackage{url}            
\usepackage{booktabs}       
\usepackage{amsfonts}
\usepackage{amsmath}
\usepackage{optidef}
\usepackage{xcolor}
\usepackage{nicefrac}       
\usepackage{microtype}      
\usepackage{float}
\usepackage{mathtools,cases}
\usepackage{empheq}
\usepackage{bm}
\usepackage[ruled,vlined]{algorithm2e}
\usepackage{pifont}
\usepackage{amsthm}
\usepackage{booktabs}
\usepackage{subfig}
\usepackage{multirow,makecell}
\usepackage[inline]{enumitem}
\usepackage{array}
\usepackage{diagbox}
\usepackage{color, colortbl}
\definecolor{Gray}{gray}{0.9}
\usepackage{pifont}

\usepackage{arydshln}
\def\eg{e.g.~}
\def\ie{i.e.~}
\def\vs{vs.~}
\def\etal{\textit{et al.~}}
\newcommand{\x}{\mathbf{x}}
\newcommand{\y}{\mathbf{y}}
\newcommand{\z}{\mathbf{z}}

\ifCLASSOPTIONcompsoc
  \usepackage[nocompress]{cite}
\else
  \usepackage{cite}
\fi
\ifCLASSINFOpdf
\else
\fi

\begin{document}
%
\title{Deep Time Series Forecasting with Shape and Temporal Criteria}

\author{Vincent~Le Guen,
        Nicolas~Thome
}

%
%

\markboth{IEEE TRANSACTIONS ON PATTERN ANALYSIS AND MACHINE INTELLIGENCE}%
{Shell \MakeLowercase{\textit{et al.}}: Deep Time Series Forecasting with Shape and Temporal Criteria}
%



\IEEEtitleabstractindextext{%
\begin{abstract}
This paper addresses the problem of multi-step time series forecasting for non-stationary signals that can present sudden changes. Current state-of-the-art deep learning forecasting methods, often trained with variants of the MSE, lack the ability to provide sharp predictions in deterministic and probabilistic contexts. To handle these challenges, we propose to incorporate shape and temporal criteria in the training objective of deep models. We define shape and temporal similarities and dissimilarities, based on a smooth relaxation of Dynamic Time Warping (DTW) and Temporal Distortion Index (TDI), that enable to build differentiable loss functions and positive semi-definite (PSD) kernels. With these tools, we introduce DILATE  (DIstortion Loss including shApe and TimE), a new objective for deterministic forecasting, that explicitly incorporates two terms supporting precise shape and temporal change detection. For probabilistic forecasting, we introduce STRIPE++ (Shape and Time diverRsIty in Probabilistic
forEcasting), a framework for providing a set of sharp and diverse forecasts, where the structured shape and time diversity is enforced with a determinantal point process (DPP) diversity loss. Extensive experiments and ablations studies on synthetic and real-world datasets confirm the benefits of leveraging shape and time features in time series forecasting.
\end{abstract}

}

\maketitle

\IEEEdisplaynontitleabstractindextext

%
\IEEEpeerreviewmaketitle

\begin{figure*}[t]
\begin{tabular}{cccc}
\includegraphics[height=4.3cm]{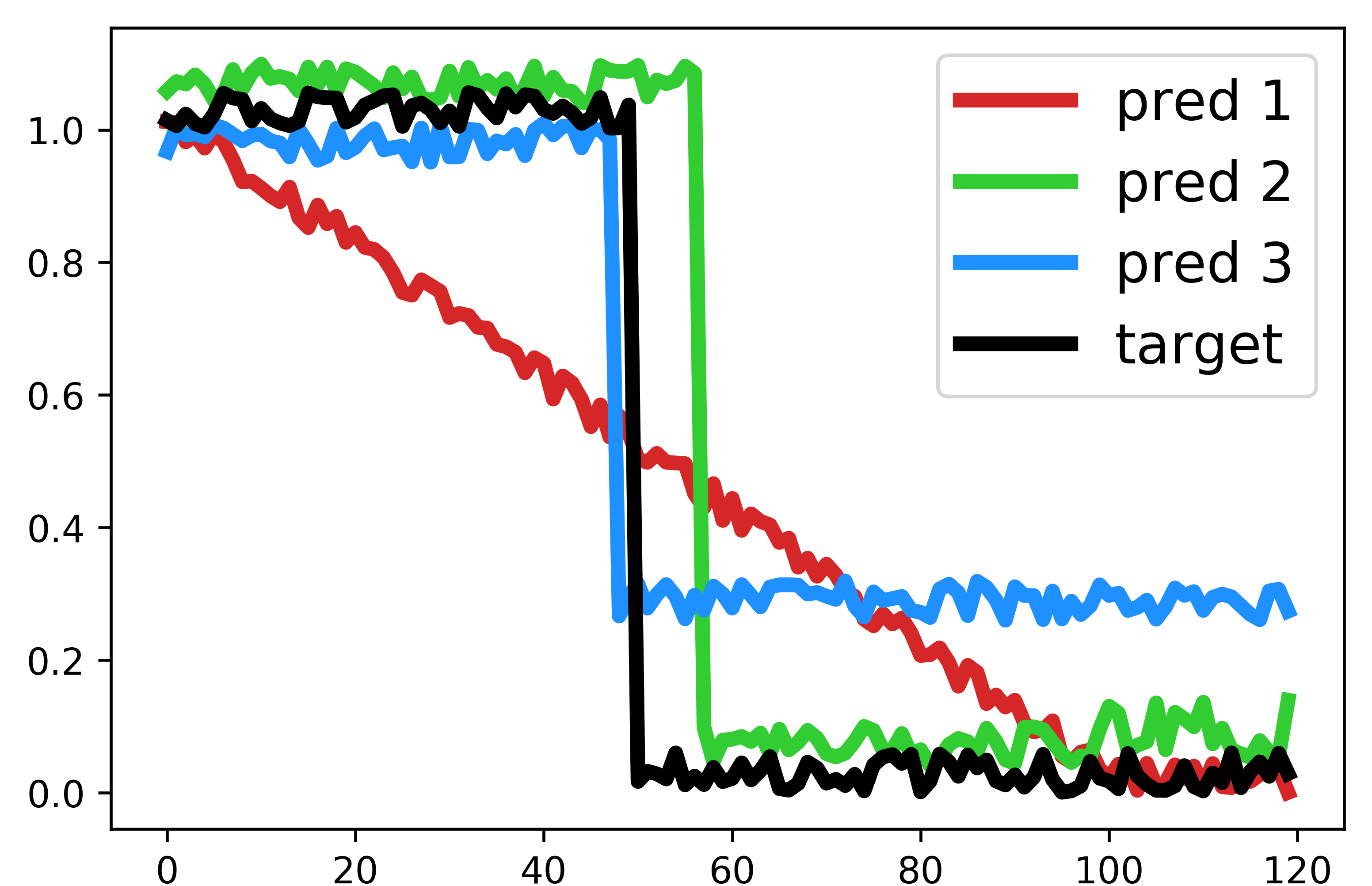}   & \hspace{-0.3cm} 
\includegraphics[height=4.3cm]{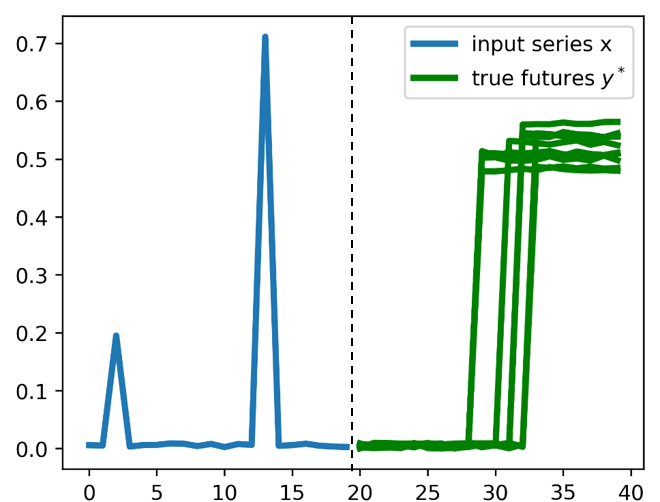}   & 
\hspace{-0.5cm} 
\includegraphics[height=4.3cm]{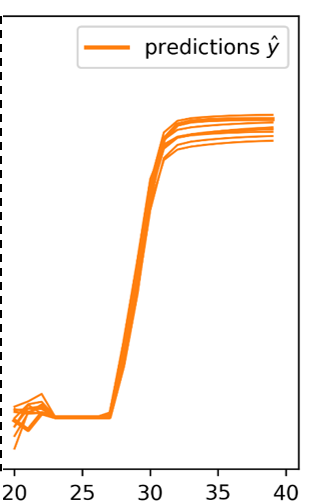}   &  
\hspace{-0.5cm} 
\includegraphics[height=4.3cm]{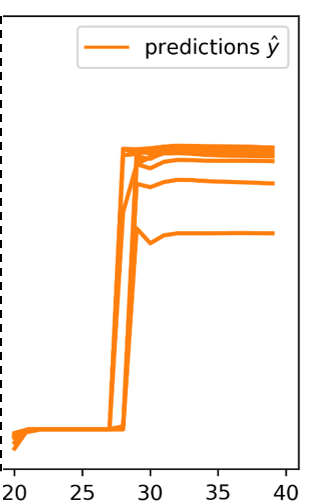}   \\
\vspace{-0.05cm}
~ &   \footnotesize{True predictive distribution}    & \hspace{-0.5cm}  \footnotesize{deep stoch. model \cite{yuan2019diverse}} & \hspace{-0.5cm}  \footnotesize{STRIPE++ (ours)}  \\
 \textbf{(a)  MSE limits in deterministic forecasting} & ~ &  \hspace{-3cm} \textbf{(b) Probabilistic forecasting} & \\
\end{tabular}{}

    \caption{(a) We illustrate the limitations of the MSE in deterministic forecasting: the three predictions (1,2,3) have the same MSE with respect to the target, contrary to our proposed DILATE loss that favours predictions 2 (correct shape, slight delay) and 3 (correct timing, inaccurate amplitude) over prediction 1. (b) In probabilistic forecasting, state-of-the-art methods trained with variants of the MSE (\eg \cite{yuan2019diverse,rasul2020multi}) loose the ability to produce sharp forecasts (middle). In contrast, our proposed STRIPE++ model ensures both sharp and diverse forecasts (right) that better match the true future distribution (left).}
    \label{fig-intro}
\end{figure*}

\IEEEraisesectionheading{\section{Introduction}\label{sec:introduction}}

Time series forecasting consists in analyzing historical signal correlations to anticipate future behaviour. Traditional 
approaches include linear autoregressive methods \cite{box2015time} or state space models \cite{durbin2012time}, which are simple yet mathematically grounded and benefit from 
interpretability. They often exploit prior knowledge based on stationarity, \eg by leveraging trend or seasonality to constrain forecasting. 

These grounding assumptions are often violated in many real-world time series that are non-stationary and can present sharp variations such as sudden drops or changes of regime. Long-term multi-step forecasting in this context is particularly challenging and arises in a wide range of important application fields, \eg analyzing traffic flows \cite{li2017diffusion,snyder2019streets}, medical records \cite{chauhan2015anomaly}, predicting sharp variations in financial markets \cite{ding2015deep} or in renewable energy production \cite{vallance2017towards,ghaderi2017deep,leguen-fisheye}, \textit{etc}.

This work focuses on forecasting multi-step future trajectories with potentially sharp variations in the deterministic and probabilistic cases. Deep neural networks are an appealing solution for this problem \cite{yu17learning,qin2017dual,lai2018modeling,salinas2017deepar,oreshkin2019n,zhou2020informer}, due to their automatic feature extraction and complex nonlinear time dependencies modeling. However, the verification criteria typically used in applications are not used at training time because they are mostly not differentiable. We may cite for instance the ramp score \cite{vallance2017towards} for assessing the detection of sharp ramping events, or the Time Distortion Index (TDI) \cite{frias2017assessing} for assessing the time delay of a particular predicted event.  

Instead, the huge majority of methods optimize at training time the Mean Squared Error (MSE) or its variants (MAE, quantile loss, \textit{etc}) as a proxy loss function. However, the MSE has important drawbacks for assessing predictions, especially in non-stationary contexts with sharp variations \cite{frias2017assessing,vallance2017towards,cuturi2017soft,florita2013identifying,wang2009mean,yang2020verification,verbois2020beyond}. We illustrate the limitations of the MSE in Figure \ref{fig-intro}. Figure \ref{fig-intro} (a) shows three deterministic predictions, which have the same MSE loss compared to the target step function (in black). Thus, the MSE does not support predictions (2) and (3) over prediction (1), although they clearly are more adequate for regulation purposes because they do anticipate the drop to come, although with a slight delay (2) or with a slightly inaccurate amplitude (3). 
~For probabilistic forecasting (Figure \ref{fig-intro} (b)), current state-of-the art probabilistic methods trained with variants of the MSE tend to produce blurry predictions that do not match the sharp steps of the true futures (in green).


In this work, we intend to bridge this train/test criterion gap by introducing shape and temporal similarities and dissimilarities for training deep forecasting models. We introduce differentiable loss functions,
~which makes them amenable to gradient-based optimization. We decouple the definition of these criteria from the introduction of general deterministic and probabilistic forecasting methods leveraging such criteria.
 

For deterministic forecasting, we introduce the DILATE training loss function (DIstortion Loss with shApe and TimE) as an alternative to the MSE. DILATE efficiently combines a shape matching term based on a smooth approximation of Dynamic Time Warping (DTW) \cite{cuturi2017soft} with a temporal term, which ensures both sharp predictions with accurate temporal localization. In Figure \ref{fig-intro} (a), DILATE favors predictions (2) and (3) over prediction (1), contrarily to the MSE.

For probabilistic forecasting, our goal is to cover the true future distribution density with a small number (\eg $N=10$) of admissible  trajectories. To this end, we introduce the STRIPE++ framework (Shape and Time diverRsIty in Probabilistic forEcasting) for producing both diverse and accurate forecasts. We leverage the proposed shape and temporal positive semi-definite (PSD) kernels to design a structured diversity measure, which is embedded into a determinantal point processes (DPP). 
~As illustrated in Figure~\ref{fig-intro} (b), STRIPE++ produces diverse predictions, which are i) representative of the ground truth future density and ii) consistent to the sharp nature of the expected forecasts.
~We validate the relevance of DILATE and STRIPE++ on extensive synthetic and real-world datasets highlighting the benefits brought up by leveraging shape and temporal features for time series forecasting.

This paper extends two previous conference papers that first introduced the DILATE loss \cite{leguen19dilate} and the STRIPE model \cite{leguen20stripe}. This submission includes the following improvements: 
\begin{itemize}
     \item  We provide a unified view of shape and temporal differentiable criteria between time series, expressed in terms of dissimilarities (loss functions) and similarities (positive semi-definite kernels). This general view enables to put into perspective our contributions for deterministic and probabilistic forecasting with respect to the literature. In that regard, the related work has substantially been extended compared to \cite{leguen19dilate,leguen20stripe}.
    \item STRIPE++ improves STRIPE \cite{leguen20stripe} at two important methodological levels for guaranteeing a good quality/diversity cooperation: we enrich the prediction model with a posterior network inspired by \cite{kolh-probunet} for better disentangling deterministic and probabilistic spaces,  
    ~and we add a quality constraint into the DPP kernels to explicitly support predictions, which are both diverse and accurate.  
    \item The experiments have been substantially extended, including new evaluations of DILATE on top of recent state-of-the-art forecasting models
    \cite{oreshkin2019n, zhou2020informer} 
    , experiments on the additional Electricity and ETTh1 datasets, a more in-depth analysis of the DTW smoothing parameter $\gamma$ and a comparison to a recent smooth DTW variant \cite{blondel2020differentiable}, and a validation of the relevance of STRIPE++ compared to STRIPE~\cite{leguen20stripe} with respect to the F1 score.

\end{itemize}

\section{Related work \label{sec:related-work}}

We review here the literature on deterministic and probabilistic time series forecasting, focusing on the multi-step non stationary context. We insist on the previous works leveraging shape and temporal features and the methods for enforcing structured diversity.

\subsection{Time series forecasting}

\textbf{Deterministic forecasting:} Traditional methods for time series forecasting are based on linear state space models \cite{durbin2012time}, including autoregressive models (\eg ARIMA \cite{box2015time}) and Exponential Smoothing \cite{hyndman2008forecasting}. These methods often exploit strong structural assumptions on data such as stationarity and seasonality that are not satisfied for many real-world time series that can present abrupt changes of distribution. Since, Recurrent Neural Networks (RNNs), in particular Long Short Term Memory Networks (LSTMs)~\cite{Hochreiter:1997:LSM:1246443.1246450}, have become popular due to their automatic feature extraction capabilities and long-term dependencies modeling. For multi-step forecasting, the most common approach is to apply recursively a one-step ahead trained model \cite{lai2018modeling}. A thorough comparison of the different multi-step strategies \cite{taieb2016bias} recommend the Direct Multi-Horizon Strategy. Of particular interest in this category are Sequence To Sequence (Seq2Seq) models  \cite{qin2017dual,fox2018deep,fan2019multi,kuznetsov2018foundations}. Recently, much effort has been devoted to design new architectures that address error accumulation in multi-step forecasting \cite{sen2019think,li2019enhancing}. In particular, Oreshkin \etal introduce the N-Beats model \cite{oreshkin2019n} with deep stacks of fully-connected layers with forward and backward residual connections. Zhou \etal propose the Informer model \cite{zhou2020informer} that leverages the self-attention Transformer architecture \cite{vaswani2017attention} to extend predictions to much farther temporal horizons.\\

\textbf{Probabilistic forecasting: } Many critical applications require forecasts associated with uncertainty to make relevant decisions. Some methods propose to estimate the variance with Monte Carlo dropout \cite{zhu2017deep,laptev2017time} or ensembling \cite{smyl2019machine}. Other methods predict the quantiles of the conditional distribution of future values \cite{wen2017multi,gasthaus2019probabilistic,wen2019deep} by minimizing the pinball loss \cite{koenker2001quantile} or the continuous ranked probability score (CRPS) \cite{gneiting2007probabilistic}. State space models, which have an inherent probabilistic nature, were recently revisited with deep learning \cite{rangapuram2018deep,kurle2020deep}. Salinas \etal introduce deepAR \cite{salinas2017deepar} by explicitly approximating the predictive distribution of the next time step with a Gaussian distribution; subsequent works investigate other parametric distributions \cite{salinas2019high}. Generative models were also used to implicitly approximate the predictive distribution, \eg with conditional variational autoencoders (cVAEs) \cite{yuan2019diverse}, conditional generative adversarial networks (cGANs) \cite{koochali2020if}, or normalizing flows \cite{rasul2020multi,de2020normalizing}. However, these methods lack the ability to produce sharp forecasts by minimizing variants of the MSE (pinball loss, Gaussian maximum likelihood), at the exception of cGANs - but which suffer from mode collapse that limits predictive diversity. Moreover, these generative models are generally represented by unstructured distributions in the latent space (\eg Gaussian), which do not allow to have an explicit control on the targeted diversity.

\subsection{Evaluation and training metrics}

Current research mainly focuses on forecasting architecture design and the question of the training loss is often overlooked. The MSE, MAE and its variants (SMAPE, \textit{etc}) are predominantly used as proxies. Many works have highlighted the limitations of the MSE to assess the ability to anticipate sharp variations
\cite{frias2017assessing,vallance2017towards,cuturi2017soft,florita2013identifying,wang2009mean,yang2020verification,verbois2020beyond}. Metrics reflecting shape and temporal localization exist but their non-differentiability makes them unsuitable for training deep models. 
For characterizing shape, Dynamic Time Warping (DTW) \cite{sakoe1990dynamic,jeong2011weighted,zhang2017dynamic} performs time series alignment, and the ramp score  \cite{florita2013identifying,vallance2017towards} assesses the detection of ramping events in wind and solar energy forecasting. Timing errors can be characterized among other ways by the Temporal Distortion Index (TDI) \cite{frias2017assessing,vallance2017towards}, or by computing detection scores (precision, recall, Hausdorff distance) after change point detection \cite{truong2019supervised}.  

 Recently, some attempts have been made to train deep neural networks based on alternatives to MSE, especially based on smooth approximations of DTW \cite{cuturi2017soft, mensch2018differentiable,abid2018learning,vayer2020time,blondel2020differentiable}. 
 The soft-DTW of Cuturi and Blondel \cite{cuturi2017soft} was further normalized to ensure a non-negative divergence \cite{blondel2020differentiable}. However, since DTW is by design invariant to elastic distortions, it completely ignores the temporal localization of the changes
 (see illustrations of soft-DTW in Figure \ref{fig:dilate_visu}). A differentiable timing error loss function based on DTW on the event (binary) space was proposed in \cite{rivest2019new} ; however it is only applicable for predicting binary time series. Some works explored the use of adversarial losses for time series \cite{yoon2019time,wu2020adversarial}, which can be seen as an implicit way of enforcing semantic criteria learned from data. However, it gives a weaker and non-interpretable control on shape and time criteria and brings additional adversarial training  difficulties. In this work, we focus on efficiently combining explicit shape and temporal differentiable criteria at training time.

\subsection{Structured diversity for prediction}

For diversifying forecasts, several repulsive schemes were studied such as the variety loss \cite{gupta2018social,thiede2019analyzing} that consists in optimizing the best sample, or entropy regularization \cite{dieng2019prescribed,wang2019nonlinear} that encourages a uniform distribution. Besides, generative models, such as variational autoencoders (VAE) \cite{kingma2013auto}, are widely used for producing multiple predictions through sampling from a latent space. However latent states are typically sampled at test time from a standard Gaussian prior distribution, resulting in an unstructured diversity. To improve this unstructured mechanism, prior works \cite{yuan2019diverse,yuan2020dlow} introduced proposal neural networks for generating the latent variables that are trained with a diversity objective.\\

\textbf{Determinantal Point Processes (DPP):} DPPs are appealing probabilistic models for describing the diversity of a set of items $\mathcal{Y}= \left\{\mathbf{y}_1,...,\mathbf{y}_N \right\}$.
A DPP is a probability distribution over all subsets of $\mathcal{Y}$ that assigns the following probability to a random subset $\mathbf{Y}$:
\begin{equation}
    \mathcal{P}_{\mathbf{K}}(\mathbf{Y}=Y) = \frac{\det(\mathbf{K}_Y)}{\sum_{Y' \subseteq \mathcal{Y}}\det(\mathbf{K}_Y')} = \frac{\det(\mathbf{K}_Y)}{\det(\mathbf{K+I})}
\end{equation}{}
where $\mathbf{K}$ is a positive semi-definite (PSD) kernel and $\mathbf{K}_A$ is its restriction to the elements indexed by $A$.  

DPPs offer efficient algorithms for sequentially sampling diverse items or maximizing the diversity of a set with a given sampling budget. Importantly, the choice of the kernel enables to incorporate prior structural knowledge on the targeted diversity. As such, DPPs have been successfully applied in various contexts, \eg document summarization \cite{gong2014diverse}, recommendation systems \cite{gillenwater2014expectation}, image generation \cite{elfeki2018gdpp} and diverse trajectory forecasting \cite{yuan2019diverse}.
 GDPP \cite{elfeki2018gdpp} proposed by Elfeki \etal is based on matching generated and true sample diversity by aligning the corresponding DPP kernels, and thus limits their use in datasets where the full distribution of possible outcomes is accessible. In contrast, our probabilistic forecasting approach is applicable in realistic scenarii where only a single future trajectory is available for each training sample. Although we share with the work of Yuan and Kitani \cite{yuan2019diverse} the goal to use DPP as diversification mechanism for future trajectories, the main limitation in~\cite{yuan2019diverse} is to use the MSE loss for training the predictor and the MSE kernel for  diversification, leading to blurred prediction, as illustrated in Figure~\ref{fig-intro} (b). In contrast, we design specific shape and time DPP kernels and we show the necessity to decouple the criteria used for quality and diversity.

\section{Differentiable shape and temporal (dis)similarities for time series }
\label{sec:criteria}

In this section, we introduce an unified view of shape and temporal criteria applicable to deep time series forecasting. We insist on their differentiability and efficient computation. We propose both a dissimilarity version useful for defining loss functions and a similarity version useful for defining positive semi-definite (PSD) kernels.

\subsection{Shape (dis)similarity}

\subsubsection{Background: Dynamic Time Warping}

To assess the shape similarity between two time series, the popular Dynamic Time Warping (DTW) method \cite{sakoe1990dynamic} seeks a minimal cost alignment for handling time distortions. 
Given two $d$-dimensional time series $\y \in \mathbb{R}^{d \times n}$ and $\z  \in \mathbb{R}^{d \times m}$ of lengths $n$ and $m$, DTW looks for an optimal warping path represented by a binary matrix $\mathbf{A}  \subset \left \{  0,1 \right \}  ^{n \times m}$ where $\mathbf{A}_{ij}=1$ if $\y_i$ is associated to $\z_j$ and 0 otherwise. The set of admissible warping paths $\mathcal{A}_{n,m}$ is composed of paths connecting the endpoints $(1,1)$ to $(n,m)$ with the following authorized moves $\rightarrow, \downarrow, \searrow$. The cost of warping path $\mathbf{A}$ is the sum of the costs along the alignment ; this cost can be written as the scalar product  $\left\langle \mathbf{A}, \mathbf{\Delta}(\y,\z) \right\rangle$, where  $\Delta(\y,\z)$  is a $n \times m$ pairwise dissimilarity matrix whose general term is typically chosen an the euclidean distance $\mathbf{\Delta}(\y,\z)_{ij} = \Vert \y_i-\z_j \Vert^2_{2}$. DTW computes the minimal cost warping path:
\begin{equation}
\text{DTW}^{\mathbf{\Delta}}(\y, \z) :=\underset{\mathbf{A} \in \mathcal{A}_{n,m}}{\min} \left \langle \mathbf{A},\mathbf{\Delta}(\y, \z) \right \rangle
\label{eq:dtw}
\end{equation}
Although the cardinality of $\mathcal{A}_{n,m}$ increases exponentially in $\min(n,m)$ \footnote{ $|\mathcal{A}_{n,m}|$ is equal to the Delannoy number $Delannoy(n,m)$ which grows exponentially in $\min(n,m)$}, DTW and the optimal path $\mathbf{A^*}$ can be computed efficiently in $\mathcal{O}(nm)$ by dynamic programming. However, a major limitation of DTW is its non-diffentiability, which prevents its integration in neural network pipelines trained with gradient-based optimization.

\subsubsection{Smooth DTW shape dissimilarity}
\label{sec:soft-dtw}
 
 For handling the non-differentiability of DTW, Cuturi and Blondel \cite{cuturi2017soft} introduced the soft-DTW by replacing the hard-minimum operator by a smooth minimum with the log-sum-exp trick $\min_{\gamma}(a_1,...,a_n) = - \gamma \log(\sum_i^n \exp(-a_i / \gamma) )$:
  \begin{multline}
 \text{DTW}^{\mathbf{\Delta}}_{\gamma}(\y, \z)  :=
 - \gamma  \log \left ( \sum_{\mathbf{A} \in \mathcal{A}_{n,m}} e ^ { - \left \langle \mathbf{A},\mathbf{\Delta}(\y, \z) \right \rangle / \gamma} \right )
\label{eq:dtwgamma}
 \end{multline}
  where $\gamma > 0$ is a smoothing parameter (when $\gamma \rightarrow 0$, this converges to the true DTW).

$\text{DTW}^{\mathbf{\Delta}}_{\gamma}$ as defined in Eq. \ref{eq:dtwgamma} is differentiable with respect to $\mathbf{\Delta}$ (and with respect to both series $\y$ and $\z$ by chain's rule, provided a differentiable cost function $\mathbf{\Delta}$).

We can interpret this relaxed DTW version by considering, instead of the unique optimal path $\mathbf{A}^*$, a Gibbs distribution over possible paths:
\begin{equation}
     p_{\gamma}(\mathbf{A} ; \mathbf{\Delta}) = \frac{1}{Z} \:  e^{- \left \langle \mathbf{A},\mathbf{\Delta}(\y, \z) \right \rangle / \gamma }
     \label{eq:gibbs}
 \end{equation}
 The soft-DTW is then the negative log-partition of this distribution: $\text{DTW}^{\mathbf{\Delta}}_{\gamma}(\y, \z)  :=  - \gamma \log Z $.

Since $\text{DTW}^{\mathbf{\Delta}}_{\gamma}(\y,\z)$ can take negative values and is not minimized for $\y=\z$, Mensch and Blondel \cite{mensch2018differentiable} normalized the soft-DTW to make it a true divergence. We found experimentally that this does not improve performances and is heavier computationally (see Appendix D.4).

\subsubsection{Shape similarity kernel}
\label{sec:shape-kernel}

Based on the soft-DTW shape dissimilarity defined in Eq. \ref{eq:dtwgamma}, we define a shape similarity kernel as follows:
\begin{equation}
    \mathcal{K}_{shape}(\y,\z) = e^{- \: \text{DTW}^{\mathbf{\Delta}}_{\gamma}(\y,\z) / \gamma}.
    \label{eq:kshape}
\end{equation}
We experiment with the following choices of kernels $\Delta_{ij} = \Delta(\y,\z)_{ij}$:
\begin{itemize}
    \item  Half-Gaussian: $\mathbf{\Delta}_{ij}=  \Vert \y_i-\z_j \Vert^2_2 +  \log (2 - e^{- \Vert \y_i-\z_j \Vert^2_2 })$
    \item L1: $\mathbf{\Delta}_{ij}= \Vert\y_i-\z_j\Vert_1$ ~~ (for $d=1$)
    \item Euclidean:  $\mathbf{\Delta}_{ij}=  \Vert \y_i-\z_j \Vert^2_2$
\end{itemize}
$\mathcal{K}_{shape}$ was proven to be positive semi-definite (PSD) for the half-Gaussian\footnote{We denote this kernel "half-Gaussian" since the corresponding $k$ kernel defined in the proof (Appendix A) equals $k(\y_i,\z_j) = e^{- \Delta(\y_i,\z_j)} = \left(\frac{1}{2} e^{-\Vert \y_i-\z_j \Vert^2})\right) \times \left(1 -  \frac{1}{2} e^{-\Vert \y_i-\z_j \Vert^2}\right)^{-1}$} and the L1 kernels \cite{cuturi2007kernel,blondel2020differentiable} and is conjectured to be PSD for the Euclidean kernel \cite{blondel2020differentiable}. Experimentally we observed that these three cost matrices lead to similar behaviour.

\begin{figure*}
    \centering
    \includegraphics[width=17cm]{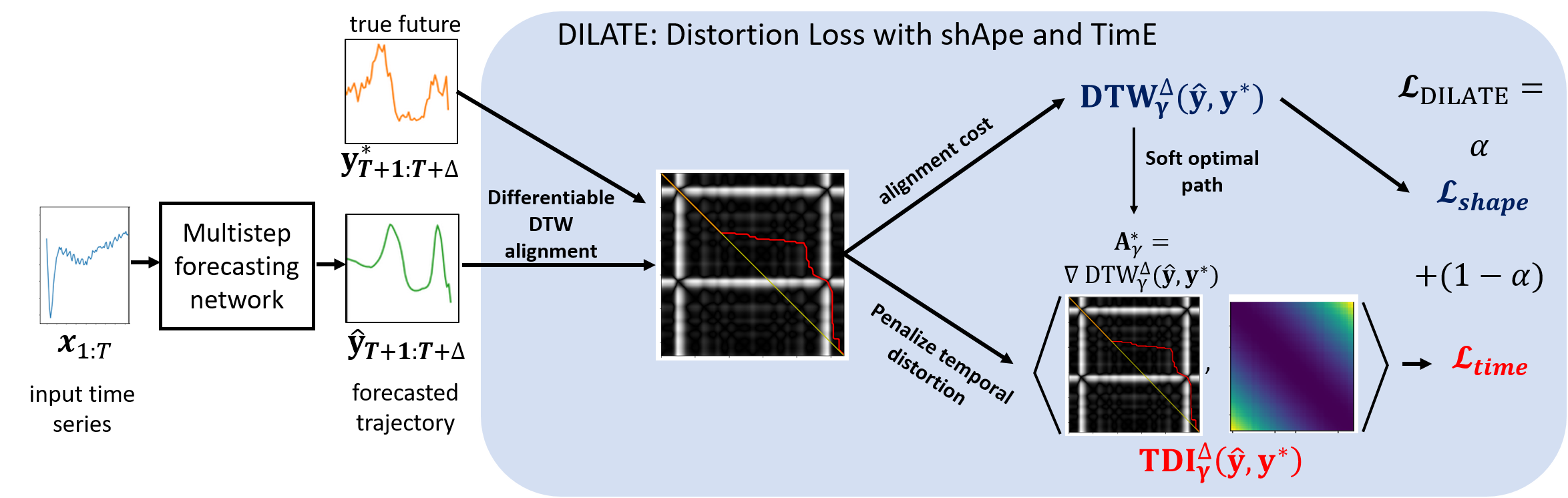}
    \caption{The DILATE loss $\mathcal{L}_{\text{DILATE}}$ for training deterministic deep time series forecasting models is composed of two terms: $\mathcal{L}_{shape}$ based on the soft DTW and $\mathcal{L}_{time}$ that penalizes the temporal distortions visible on the soft optimal path. The overall loss  $\mathcal{L}_{\text{DILATE}}$ is differentiable, 
    and we provide an efficient implementation of its forward and backward passes.}
    \label{fig:dilate}
\end{figure*}

\subsection{Temporal (dis)similarity}

Quantifying the temporal similarity between two time series consists in analyzing the time delays between matched patterns detected in both series. As discussed in introduction, it of great importance for many applications to anticipate sharp variations. 

\subsubsection{Smooth temporal distortion index}

A common temporal similarity is the Temporal Distortion Index (TDI) \cite{frias2017assessing, vallance2017towards}. The TDI computes the approximate area included between the optimal path $\mathbf{A^*}$ and the first diagonal, characterizing the presence of temporal distortion. A generalized version of the TDI, that we proposed in \cite{leguen19dilate}, can be written:
\begin{equation}
 \text{TDI}^{\mathbf{\Delta, \Omega_{dissim}}}(\y, \z) :=   \langle  \mathbf{A}^*, \mathbf{\Omega_{dissim}}  \rangle \>  
 \label{eq:tdi}
\end{equation}
where $ \mathbf{A}^* =  \underset{\mathbf{A} \in \mathcal{A}_{n,m}}{\arg \min} \left \langle \mathbf{A},\mathbf{\Delta}(\y, \z) \right \rangle$ is the DTW optimal path and  $\mathbf{\Omega_{dissim}} \in \mathbb{R}^{n \times m}$ is a matrix penalizing the association between $\y_i$ and $\z_{j}$ for $i \neq j$. We typically choose a quadratic penalization  $\mathbf{\Omega_{dissim}}(i,j) \propto (i-j)^2$, but other variants can encode prior knowledge and penalize more heavily late than early predictions, and \textit{vice-versa}.

The TDI dissimilarity defined in Eq. \ref{eq:tdi} is however non-differentiable, since the optimal path $\mathbf{A}^*$ is not differentiable with respect to $\mathbf{\Delta}$. We handle this problem
by defining a relaxed optimal path $\mathbf{A}^*_{\gamma}$ as the gradient of $\text{DTW}_{\gamma}^{\mathbf{\Delta}}$: 
\begin{align}
    \mathbf{A}^*_{\gamma}  := \nabla_{\mathbf{\Delta}} \text{DTW}^{\mathbf{\Delta}}_{\gamma}(\y, \z) =  \frac{1}{Z}  \sum_{\mathbf{A} \in \mathcal{A}_{n,m}} \mathbf{A} \:  e^{- \left \langle \mathbf{A},\mathbf{\Delta}(\y, \z) \right \rangle / \gamma } 
 \label{eq:grad_dtw}
\end{align}
The expression in Eq. \ref{eq:grad_dtw} results from a direct computation from Eq. \ref{eq:dtwgamma}. Notice that this soft optimal path corresponds to the expected path $\mathbf{A}^*_{\gamma} = \mathbb{E}_{p_{\gamma}(\cdot ; \mathbf{\Delta})} [\mathbf{A}]$ under the Gibbs distribution in Eq. \ref{eq:gibbs}. Note also that $\mathbf{A}^*_{\gamma}$ becomes a soft assignment, \ie $\mathbf{A}^*_{\gamma}(i,j)$ represents the probability for a path to contain the cell $(i,j)$. An illustration of soft optimal paths with the influence of $\gamma$ is given in Figure \ref{fig:influ_gamma}.

We can now define a differentiable version of the TDI:
\begin{multline}
    \text{TDI}^{\mathbf{\Delta,\Omega_{dissim}}}_{\gamma}(\y,\z) := \left \langle  \mathbf{A}_{\gamma}^* , \mathbf{\Omega_{dissim}}  \right \rangle  = \\
     \dfrac{1}{Z}  \sum_{\mathbf{A} \in \mathcal{A}_{n,m}}  \left \langle  \mathbf{A}, \mathbf{\Omega_{dissim}} \right \rangle  e^{-\frac{ \left \langle \mathbf{A},\mathbf{\Delta}(\y, \z) \right \rangle}{\gamma} } 
     \label{eq:temporal}
\end{multline}
which corresponds to the expected value of the TDI under the Gibbs distribution.

\subsubsection{Temporal similarity kernel}

Based on the temporal dissimilarity in Eq. \ref{eq:temporal} and the shape similarity kernel in Eq. \ref{eq:kshape}, we can define a time similarity as follows:
\begin{equation}
    \mathcal{K}_{time}(\y,\z) := e^{- \text{DTW}^{\mathbf{\Delta}}_{\gamma}(\y,\z) / \gamma}
  \times \text{TDI}^{\mathbf{\Delta, {\Omega_{sim}}}}_{\gamma} (\y,\z)
  \label{eq:Ktime}
\end{equation}
where in this case, we use a similarity matrix $\mathbf{\Omega_{sim}}$ favoring pairs of time series with low temporal distortion, \ie with an optimal path near the main diagonal. We typically choose a pointwise inverse of $\mathbf{\Omega_{dissim}}$: $\mathbf{\Omega_{sim}}(i,j) = \frac{1}{(i-j)^2+1}$. We prove that $ \mathcal{K}_{time}$ defines a valid PSD temporal kernel (proof in Appendix A). \\

The table below provides an overview of the shape and temporal criteria introduced in this work:\\

\begin{tabular}{c|c|c}
  criterion    & differentiable loss   & PSD sim. kernel  \\ 
  \hline
    shape & $\text{DTW}^{\mathbf{\Delta}}_{\gamma}(\y, \z)$   &  $ e^{-  \: \text{DTW}^{\mathbf{\Delta}}_{\gamma}(\y,\z) / \gamma} $  \\ 
    \hline
    time &   $\text{TDI}^{\mathbf{\Delta,\Omega_{dissim}}}_{\gamma}(\y,\z)$  & 
    \begin{tabular}{c}
       $e^{-  \text{DTW}^{\mathbf{\Delta}}_{\gamma}(\y,\z)/ \gamma} $    \\
           $  \times  \text{TDI}^{\mathbf{\Delta, {\Omega_{sim}}}}_{\gamma} (\y,\z)$
    \end{tabular}

\end{tabular}

\vspace{0.2cm}
\textbf{Efficient forward and backward computation:} The direct computation of the shape and temporal dissimilarities in Eq. \ref{eq:dtwgamma} and Eq. \ref{eq:temporal} is intractable, due to the exponential growth of cardinal of $\mathcal{A}_{n,m}$. We provide a careful implementation of the forward and backward passes in order to make learning efficient, limiting the forward and backward complexity to $\mathcal{O}(nm)$ (see details in Appendix B).

\subsubsection{Discussion on differentiability} One may mask why not directly optimizing the true DTW and TDI objectives with subgradient methods instead of deriving smooth surrogates? The DTW is indeed continuous with respect to both input time series and when the optimal warping path  $\textbf{A}^*$ is unique (which is almost always the case for continuous data), DTW is differentiable almost everywhere with gradient $\textbf{A}^*$. However, a first difficulty of directly optimizing DTW is that the gradient is discontinuous at the points where a change in the input time series causes a change in the optimal path  $\textbf{A}^*$, which can hamper the performances of gradient-based methods. A second difficulty is that the DTW objective is non-convex, making the subgradient method prone to collapsing to local minima \cite{cuturi2017soft}. Furthermore, for defining the TDI in the DILATE loss, we need second-order differentiability, which exacerbates the aforementioned difficulties.

Considering a smooth relaxation of DTW and TDI alleviates both difficulties. First, these relaxations are continuously differentiable by design. Second, the smoothing convexifies the DTW and TDI objectives and provide a better optimization landscape. This point was noted by several works \cite{cuturi2017soft,mensch2018differentiable} which showed that the soft-DTW yielded better performances than the DTW optimized with subgradient, even when evaluated with the true DTW. We have also confirmed this experimentally for our smooth shape and time surrogate criteria in section \ref{sec:dilate-analysis}.

\section{Application to time series forecasting}

 Given an input sequence $\x_{1:T}=(\x_1,\dots,\x_T) \in \mathbb{R}^{p \times T}$, we consider as discussed in introduction the multi-step time series forecasting problem in two important contexts:  \textbf{(1) deterministic forecasting} consisting in predicting a $\tau$-steps future trajectory  $ \hat{\y} = (\hat{\y}_{T+1},\dots, \hat{\y}_{T+\tau} )  \in  \mathbb{R}^{d \times \tau} $ and \textbf{(2) probabilistic forecasting} where we seek to predict a set of $N$ future trajectories  $ \{ \hat{\y}^{(i)} \}_{i=1..N}  \in  \mathbb{R}^{d \times \tau}$ (corresponding to diverse scenarii sampled from the true future distribution $\hat{\y}^{(i)} \sim p(\cdot |\x_{1:T})$). 
 
 In this section, we introduce two general methods to leverage differentiable shape and temporal criteria - such as those derived in section \ref{sec:criteria} - for deterministic and probabilistic forecasting. 
 We introduce in section \ref{sec:DILATE} a differentiable loss function, termed DILATE, for tackling the deterministic forecasting problem (1) in non-stationary contexts, that ensures sharp predictions with accurate temporal localization. In section \ref{sec:stripe}, we extend these ideas to the probabilistic problem (2) with the STRIPE++ model that enforces structured shape and temporal diversity with determinantal point processes (DPP).

\subsection{DILATE loss for deterministic forecasting}
\label{sec:DILATE}

Our proposed framework for deterministic forecasting is depicted in Figure \ref{fig:dilate}. We introduce the DIstortion Loss with shApe and TimE (DILATE) for training any deterministic deep multi-step forecasting model. Crucially, the DILATE loss needs to be differentiable in order to train models with gradient-based optimization.

The DILATE objective function, which compares the prediction $\hat{\y} =  (\hat{\y}_{T+1},\dots, \hat{\y}_{T+\tau} )$ with the actual ground truth future trajectory $\y^* = (\y^*_{T+1},\dots,\y^*_{T+\tau})$, is composed of two terms balanced by the hyperparameter $\alpha \in [0,1]$:
\begin{multline}
\label{eq:stdl}
\mathcal{L}_{\text{DILATE}}(\hat{\y}, \y^*) = \alpha~\mathcal{L}_{shape}(\hat{\y}, \y^*) + (1-\alpha)~ \mathcal{L}_{time}(\hat{\y}, \y^*)\\
 = \alpha ~\text{DTW}^{\mathbf{\Delta}}_{\gamma}(\hat{\y}, \y^*) + (1-\alpha)~ \text{TDI}^{\mathbf{\Delta},\mathbf{\Omega_{dissim}}}_{\gamma}(\hat{\y}, \y^*)
\end{multline}

\textbf{Discussion:} A variant of our approach to combine shape and temporal penalization would be to incorporate a temporal term inside our smooth $\mathcal{L}_{shape}$ function in Eq. \ref{eq:dtwgamma}, \ie: 
\begin{multline}
\mathcal{L}_{\text{DILATE}^t}(\hat{\y}_i, \y^*_{i}) := \\ - \gamma  \log \left ( \sum_{\mathbf{A} \in \mathcal{A}_{n,m}} \exp\left ( - \textstyle \frac{ \left \langle \mathbf{A} , \alpha \mathbf{\Delta}(\hat{\y}_i, \y^*_{i}) + (1-\alpha) \mathbf{\Omega} \right \rangle}{\gamma} \right ) \right ) 
\label{eq:smoothwdtw}
\end{multline}

We can notice that Eq. \ref{eq:smoothwdtw} reduces to minimizing $\left \langle \mathbf{A} , \alpha \mathbf{\Delta}(\hat{\y}_i, \y^*_{i}) + (1-\alpha) \mathbf{\Omega}  \right \rangle$ when $\gamma \to 0^+$. In this case, $\mathcal{L}_{\text{DILATE}^t}$ can recover DTW variants studied in the literature to bias the computation based on penalizing sequence misalignment, by designing specific $\mathbf{\Omega}$ matrices:
\begin{center}
 \begin{adjustbox}{max width=\linewidth}  
  \begin{tabular}{c|c}
  
  \begin{tabular}{c}
     Sakoe-Chiba DTW \\
      band constraint \cite{sakoe1990dynamic}    
  \end{tabular}

  & $\Omega(i,j) =$ 
  $\begin{cases}
         + \infty  \text{~if~} |i-j|>T       \\
       0 \text{~~ otherwise}    
    \end{cases}$
  \\ \hline
  Weighted DTW \cite{jeong2011weighted}   & $\Omega(i,j) = f(|i-j|)$ \text{~ $f \nearrow$}
\end{tabular}  
\end{adjustbox}
\end{center}

$\mathcal{L}_{\text{DILATE}^t}$ in Eq. \ref{eq:smoothwdtw} enables to train deep neural networks with a smooth loss combining shape and temporal criteria. However, $\mathcal{L}_{\text{DILATE}^t}$ presents limited capacities for disentangling the shape and temporal errors, since the optimal path is computed from both shape and temporal terms. In contrast, our $\mathcal{L}_{\text{DILATE}}$ loss in Eq. \ref{eq:stdl} separates the loss into two shape and temporal  
components, the temporal penalization being applied to the optimal unconstrained DTW path.

\begin{figure*}
    \centering
    \includegraphics[width=18cm]{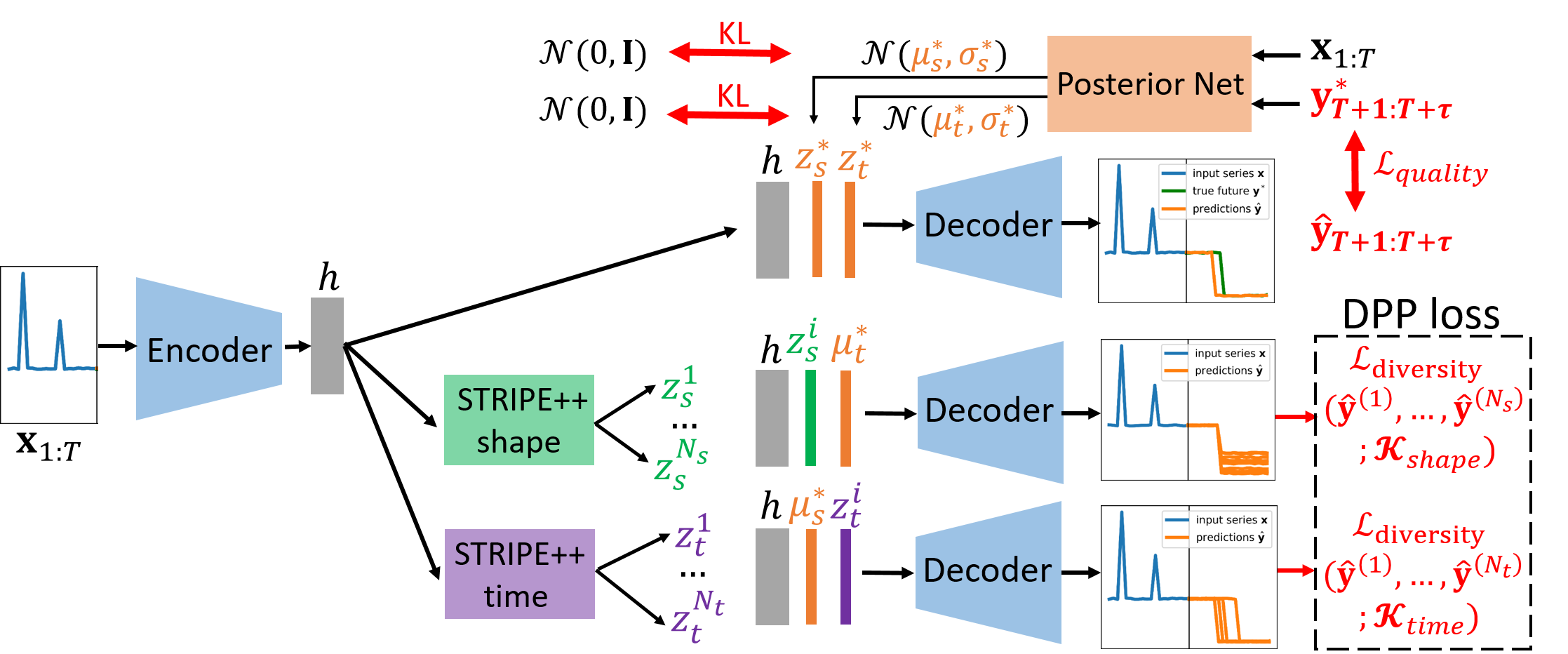}
    \caption{Our STRIPE++ model builds upon a forecasting architecture trained with a quality loss $\mathcal{L}_{quality}$ enforcing sharp predictions.
    The latent state is disentangled into a deterministic part $h$ from the encoder and two stochastic codes $z_s$ and $z_t$ that account for the shape and time variations. First step (figure upper part), we train the predictor with a quality loss, the stochastic codes are sampled from a posterior network. Second step (bottom part), we diversify the predictions with two STRIPE++ shape and time proposal networks trained with a DPP diversity loss (keeping the encoder and decoder frozen).}
    \label{fig:stripe}
\end{figure*}

\subsection{Probabilistic forecasting with structured diversity}
\label{sec:stripe}

We introduce the STRIPE++ framework (Shape and Time diverRsIty in Probabilistic
forEcasting), that extends STRIPE \cite{leguen20stripe}. Depicted in Figure \ref{fig:stripe}, STRIPE++ builds upon a general multi-step forecasting pipeline: the input time series $\x_{1:T}$ is fed into an encoder that summarizes the input into a latent vector $h$. This context vector $h$ is then transformed by a decoder into a future trajectory. 

The key idea of STRIPE++ is to augment the deterministic latent state $h$ with stochastic diversifying variables $z_s$ (resp. $z_t$) meant to capture the shape (resp. temporal) variations of the future time series. We distinguish two phases for training the overall model: (i) we train the predictor with a quality loss and (ii) we train the diversifying STRIPE++ mechanism with a DPP diversity loss (with the weights of the predictor frozen). For both of these steps, we detail now how the diversifying variables are sampled.

\subsubsection{Training the predictor with a quality loss}
For training the predictor (upper part in Figure \ref{fig:stripe}) with possibly multiple admissible futures as supervision, we take inspiration from the probabilistic U-Net \cite{kolh-probunet} and introduce a posterior network from which to sample the diversifying variables $z_s^*$ and $z_t^*$ (which represent the shape and temporal variant attached to a particular future $\y^*$). The posterior net outputs the parameters $\mu_s^*$ and $\sigma_s^*$ of a  Gaussian distribution  $\mathcal{N}(\mu_s^*,\sigma_s^*)$ for parameterizing  the shape posterior distribution  $q(z_s | \x,\y^*)$ (and similarly for the temporal posterior distribution).

To train this generative model (encoder, decoder and posterior networks), we resort to variational inference \cite{kingma2013auto} and maximize the evidence lower bound (ELBO) of the log-likelihood, or equivalently, minimize the following prediction loss over all training examples:
\begin{multline}
    \mathcal{L}_{prediction}(\hat{\y},\y^*) = \mathcal{L}_{quality}(\hat{\y},\y^*) \; + \\  \text{KL} \left( q(z_s | \x,\y^*)\;||\;p(z_s) \right) + \text{KL}\left( q(z_t | \x,\y^*)\;||\;p(z_t) \right)
\end{multline}
In our non-stationary context, we choose the DILATE loss for $\mathcal{L}_{quality}$, in order to guarantee sharp predictions with accurate temporal localization. The Kullback-Leibler (KL) losses enforce that the shape posterior distribution $q(z_s | \x,\y^*)$ matches a prior distribution $p(z_s)$ (we use a Gaussian prior $\mathcal{N}(0,\mathbf{I})$ - a common choice in variational inference).

\subsubsection{Training the STRIPE++ diversification mechanism}

For including structured shape and temporal diversity (lower part in Figure \ref{fig:stripe}), we introduce two proposal neural networks STRIPE$^{++}_{\text{shape}}$ and STRIPE$^{++}_{\text{time}}$ that aim to produce a set of $N_s$ shape latent codes $\left\{z_s^i\right\}_{{i=1..N_s}} \in \mathbb{R}^k$ (resp. $N_t$ time codes  $\left\{z_t^i\right\}_{{i=1..N_t}} \in \mathbb{R}^k$) dedicated to generate diverse trajectories in terms of shape (resp. time). 

When training STRIPE$^{++}_{\text{shape}}$ (the description for STRIPE$^{++}_{\text{time}}$ is similar), we concatenate $h$ with the posterior time latent code $\mu_t^*$ and the $N_s$ shape latent codes $z_s^i$ provided by STRIPE$^{++}_{\text{shape}}$, which leads to $N_s$ future trajectories $\hat{\y}^{i} = \text{Decoder}\left( (h, z_s^i , \mu_t^*) \right)$, $i=1..N_s$\footnote{If there exists multiple futures as supervision, we repeat this operation for each posterior latent code $\mu_t^{*,j}$ (it corresponds to consider each tuple $(\x_{1:T},\y^{*,j})$ as a separate training example).}. The shape diversity of this set of $N_s$ trajectories is then enforced by a shape diversity loss that we describe below.\\

\textbf{DPP diversity loss:} We resort to determinantal point processes (DPP) for their appealing properties for maximizing the diversity of a set of items $\mathcal{Y} = \left\{  \y_1,...,\y_N \right\}$ given a fixed sampling budget $N$ and for structuring diversity via the choice of the DPP kernel. Following \cite{yuan2019diverse}, we minimize the negative expected cardinality of a random subset $Y$ from the DPP:
\begin{align}
    \mathcal{L}_{diversity}(\mathcal{Y} ; \mathbf{K}) &= -\mathbb{E}_{Y \sim \text{DPP}(\mathcal{K})} |Y| \\ &= - Tr(\mathbf{I}-(\mathbf{K}+\mathbf{I})^{-1})
    \label{eq:ldiversity}
\end{align}{}
Intuitively, a larger expected cardinality means a more diverse sampled set according to kernel $\mathcal{K}$. We provide more details on DPPs and the derivation of $ \mathcal{L}_{diversity}$ in Appendix E.1. This loss is differentiable and can be computed in closed form.\\

\textbf{Quality regularizer in the DPP:} When training the shape and time proposal networks with the diversity loss, we do not have control over the quality of predictions, which can deteriorate to improve diversity. To address this, we introduce a quality regularization term in the DPP kernels. Crucially, we decouple the criteria used for quality (DILATE) and diversity (shape or time). $\mathcal{K}_{shape}$ maximizes the shape (DTW) diversity, while maintaining a globally low DILATE loss (thus playing on the temporal localization to ensure a good tradeoff). This contrasts with \cite{yuan2019diverse} which uses the same MSE criterion for both quality and diversity (see Appendix E.3 for a detailed explanation). In practice,  we introduce a quality vector $\mathbf{q}= (q_1,\dots,q_{N_s})$ between the prediction $\hat{\y}^i$ and the ground truth $\y^*$ \footnote{If there are multiple futures as supervision, we again consider each tuple (input sequence, possible future) as a separate training example.}. We choose $q_i = \mu  (1 - \text{DILATE}(\hat{\y}^i, \y^*))$, where $\mu > 0$ is a hyperparameter to tune the influence of the quality regularization. The modified shape kernel becomes (and similarly for the time kernel):
\begin{equation}
\Tilde{\textbf{K}}_{shape} = \text{Diag}(\textbf{q}) ~ \textbf{K}_{shape} ~ \text{Diag}(\textbf{q})
\label{eq:kshape-tilde}
\end{equation}
This decomposition enables to sample sets of items of both high quality and diversity:
\begin{equation}
    \mathcal{P}_{\mathbf{\Tilde{K}}}(\mathbf{Y}=Y) \propto \left( \prod_{i \in Y} q_i^2  \right) \det(\mathbf{K}_Y)
\end{equation}{}

 We then train STRIPE$^{++}_{\text{shape}}$ by applying the shape kernel $\Tilde{\textbf{K}}_{shape}$  (Eq. \ref{eq:kshape-tilde}) to the set of $N_s$ shape future trajectories $\mathcal{L}_{diversity}(\hat{\y}^{1},\dots,\hat{\y}^{N_s} ; \Tilde{\mathbf{K}}_{shape})$ and STRIPE$^{++}_{\text{time}}$ by applying the  time kernel $\Tilde{\textbf{K}}_{time}$ to the set of $N_t$ time future trajectories $\mathcal{L}_{diversity}(\hat{\y}^{1},\dots,\hat{\y}^{N_t} ; \Tilde{\mathbf{K}}_{time})$.

 \subsubsection{Diverse trajectory generation at test time}
 
 At test time, the posterior network is discarded and we only rely on the trained encoder, STRIPE$^{++}_{\text{shape}}$, STRIPE$^{++}_{\text{time}}$ proposal networks and decoder to generate future predictions. More precisely, we combine the shape and temporal proposals $\left\{ z_s^i \right\}_{i=1..N_s}$ and $\left\{ z_t^j \right\}_{j=1..N_t}$  to obtain $N_s \times N_t$ predictions $\hat{\y}^{i,j} =  \text{Decoder}((h,z_s^i,z_t^j))$.

\subsubsection{Discussion: differences to prior work STRIPE}
\label{sec:diff_stripe}


Our proposed STRIPE++ model improves over STRIPE \cite{leguen20stripe} by gaining a better control on quality at the diversification stage. This leads to forecasts of better quality  while maintaining the diversity performances (see the experiments in Table \ref{tab:stripe}).
This is attained with two mechanisms described above:
\begin{itemize}
\item the use of a posterior network which provides the diversification variables $z_s^*$ and $z_t^*$ attached to a particular true future trajectory $\y^*$. This ensures a better disentanglement between the deterministic latent variable $h$ and the stochastic diversification variable $z_s$ and $z_t$. In contrast in STRIPE \cite{leguen20stripe}, the $z_t$ variables are random standard Gaussian samples when training STRIPE-shape.
 
    \item the quality constraint in the diversity kernels $\Tilde{\textbf{K}}_{shape}$ and $\Tilde{\textbf{K}}_{time}$ which enables to favour predictions of both good quality and diversity.
\end{itemize}

Note that an important additional benefit of using a posterior net at training time is the ability to accommodate multiple future supervision trajectories, whereas in \cite{leguen20stripe} the diversifying variables are zeroed when training the predictor, meaning that the future ambiguity would be captured in $h$, and not in $z_s$ and $z_t$.

Another minor difference with STRIPE \cite{leguen20stripe} is that we discarded the conditioning of STRIPE-time on the shape codes generated by STRIPE-shape (which leads to no experimental benefits over a non-conditional proposal network).

\section{Experiments}

We evaluate our approach on the two tasks presented in introduction: deterministic and probabilistic time series forecasting, focusing particularly on non-stationary series that can present sudden changes.

\begin{table*}[t]
      \caption{\textbf{DILATE forecasting results on generic MLP and RNN architectures}, averaged over 10 runs (mean $\pm$ standard deviation). Metrics are scaled for readability. For each experiment, best method(s) (Student t-test) in bold.}    
      \centering
    \begin{tabular}{llccc|ccc}
    \toprule
     \multicolumn{2}{c}{}  &  \multicolumn{3}{c|}{\textbf{Fully connected network (MLP)}}       & \multicolumn{3}{c}{\textbf{Recurrent neural network (Seq2Seq)}}    \\
     \hline
       Dataset      &  \diagbox{Eval}{Train}   &  MSE &   $\text{DTW}_{\gamma}^{\mathbf{\Delta}}$~\cite{cuturi2017soft} &  DILATE (ours)  &  MSE  &   $\text{DTW}_{\gamma}^{\mathbf{\Delta}}$~\cite{cuturi2017soft} &  DILATE (ours) \\ 
       \hline
    ~         & MSE (x1000)  & ~  \textbf{16.5 $\pm$ 1.4}     & ~   48.2 $\pm$  4.0      & ~   \textbf{16.7$\pm$  1.8}        & ~         \textbf{11.0 $\pm$  1.7}        & ~  23.1 $\pm$  4.5              & ~  \textbf{12.1 $\pm$  1.3}          \\
    Synthetic & DTW (x10)     & ~   38.6 $\pm$ 1.28          & ~  \textbf{27.3 $\pm$ 1.37}          & ~  32.1 $\pm$ 5.33       & ~   \textbf{24.6 $\pm$ 1.20}                & ~  \textbf{22.7 $\pm$ 3.55}             & ~  \textbf{23.1 $\pm$ 2.44}           \\
    ~         & TDI (x10)     & ~  15.3 $\pm$ 1.39           & ~  26.9 $\pm$ 4.16           & ~   \textbf{13.8 $\pm$ 0.71}       & ~    17.2 $\pm$ 1.22              & ~      20.0 $\pm$ 3.72          & ~   \textbf{14.8 $\pm$ 1.29}          \\ 
~    & Ramp (x10)  & 5.21 $\pm$ 0.10  &  \textbf{2.04 $\pm$ 0.23}  & 3.41 $\pm$ 0.29  &  5.80 $\pm$ 0.10 & \textbf{4.27 $\pm$ 0.8} & 4.99 $\pm$ 0.46   \\
~ & Hausdorff (x1) & 4.04 $\pm$ 0.28 & 4.71 $\pm$ 0.50  & \textbf{3.71 $\pm$ 0.12}  & 2.87 $\pm$ 0.13 & 3.45 $\pm$ 0.32 & \textbf{2.70 $\pm$ 0.17}  \\
\midrule
     ~         & MSE (x100) & ~  \textbf{31.5 $\pm$ 1.39}   & ~    70.9 $\pm$ 37.2  & ~    37.2 $\pm$ 3.59    & ~    \textbf{21.2 $\pm$ 2.24}    & ~   75.1 $\pm$ 6.30              & ~      30.3 $\pm$ 4.10             \\
    ECG       & DTW (x10)     & ~     19.5 $\pm$ 0.16          & ~    18.4 $\pm$ 0.75         & ~   \textbf{17.7 $\pm$ 0.43}       & ~     17.8 $\pm$ 1.62              & ~      17.1 $\pm$ 0.65           & ~  \textbf{16.1 $\pm$ 0.16}            \\
    ~         & TDI (x10)     & ~     \textbf{7.58 $\pm$ 0.19}          & ~    17.9 $\pm$ 0.7      & ~   \textbf{7.21 $\pm$ 0.89}       & ~    8.27 $\pm$ 1.03              & ~   27.2 $\pm$ 11.1              & ~    \textbf{6.59 $\pm$ 0.79}             \\
    ~ & Ramp (x1) & \textbf{4.9 $\pm$ 0.1}  & 5.1 $\pm$ 0.3  &  \textbf{5.0 $\pm$ 0.1} &    \textbf{4.84 $\pm$ 0.24}              & ~      \textbf{4.79 $\pm$ 0.37}           & ~  \textbf{4.80 $\pm$ 0.25}    \\
    ~ & Hausdorff (x1) & \textbf{4.1 $\pm$ 0.1} & 6.3 $\pm$ 0.6  & 4.7 $\pm$ 0.3  &    \textbf{4.32 $\pm$ 0.51}     & ~   6.16 $\pm$ 0.85               & ~ \textbf{4.23 $\pm$ 0.41}        \\
 \midrule
    ~         & MSE (x1000) & ~    \textbf{6.58 $\pm$ 0.11}           & ~   25.2 $\pm$ 2.3          & ~   19.3 $\pm$ 0.80       & ~        \textbf{8.90 $\pm$ 1.1}           & ~      22.2 $\pm$ 2.6           & ~     \textbf{10.0 $\pm$ 2.6}         \\
    Traffic   & DTW (x100)     & ~   25.2 $\pm$ 0.17            & ~    \textbf{23.4 $\pm$ 5.40}         & ~   \textbf{23.1 $\pm$ 0.41}   & ~    24.6 $\pm$ 1.85              & ~      \textbf{22.6 $\pm$ 1.34}           & ~    \textbf{23.0 $\pm$ 1.62}          \\
    ~         & TDI (x100)     & ~   24.8 $\pm$ 1.1            & ~    27.4 $\pm$ 5.01         & ~  \textbf{16.7 $\pm$ 0.51}        & ~     \textbf{15.4 $\pm$ 2.25}              & ~     22.3 $\pm$ 3.66            & ~    \textbf{14.4$\pm$  1.58}          \\
    ~ & Ramp (x10) & 6.18 $\pm$ 0.1  & \textbf{5.59 $\pm$ 0.1}  &  \textbf{5.6 $\pm$ 0.1}   &  6.29 $\pm$ 0.32             & ~     \textbf{5.78 $\pm$ 0.41}           & ~    \textbf{5.93 $\pm$ 0.24}    \\
    ~ & Hausdorff (x1) & \textbf{1.99 $\pm$ 0.2}  & \textbf{1.91 $\pm$ 0.3} &  \textbf{1.94 $\pm$ 0.2}  &   \textbf{2.16 $\pm$ 0.38}           & ~      \textbf{2.29 $\pm$ 0.33}          & ~     \textbf{2.13 $\pm$ 0.51}   \\ 
\bottomrule
    \end{tabular}
    \label{results1}  
\end{table*}

\begin{table*}
    \caption{\textbf{DILATE forecasting results on state-of-the-art architectures N-Beats \cite{oreshkin2019n} and Informer \cite{zhou2020informer}}. Evaluation metrics are scaled for readability. Results are averaged over 10 runs, best(s) method(s) in bold (Student t-test).}
    \centering
    \begin{tabular}{cccccccc}
    \toprule
      Dataset & Model & MSE  & DTW & Ramp & TDI & Hausdorff & DILATE    \\
      \midrule
Synthetic   & N-Beats \cite{oreshkin2019n} MSE & \textbf{13.6 $\pm$ 0.5} & 24.9 $\pm$ 0.6 & 5.9 $\pm$ 0.1 & \textbf{13.8 $\pm$ 1.1} & \textbf{2.8 $\pm$ 0.1} & \textbf{19.3 $\pm$ 0.5}  \\
 & N-Beats \cite{oreshkin2019n} DILATE & \textbf{13.3 $\pm$ 0.7} & \textbf{23.4 $\pm$ 0.8}  & \textbf{4.8 $\pm$ 0.4}  &  \textbf{14.4 $\pm$ 1.3}  & \textbf{2.7 $\pm$ 0.5}  & \textbf{18.9 $\pm$ 0.8}  \\
  \cdashline{2-8}
   & Informer \cite{zhou2020informer} MSE & \textbf{10.4 $\pm$ 0.3} & 20.1 $\pm$ 1.1  & 4.3 $\pm$ 0.3  & 13.1 $\pm$ 0.9  & \textbf{2.5 $\pm$ 0.1} & 16.6 $\pm$ 0.8   \\
 & Informer \cite{zhou2020informer} DILATE & 11.8 $\pm$ 0.7  &  \textbf{18.5 $\pm$ 1.2}  & \textbf{2.4 $\pm$ 0.3}  & \textbf{11.6 $\pm$ 0.9}  & \textbf{2.4 $\pm$ 0.9} & \textbf{15.1 $\pm$ 0.7} \\
  \midrule
 Electricity  & N-Beats \cite{oreshkin2019n} MSE & \textbf{24.8 $\pm$ 0.4} & \textbf{15.6 $\pm$ 0.2} & \textbf{13.3  $\pm$ 0.3} & 4.6 $\pm$ 0.1 & \textbf{2.6 $\pm$ 0.3}  & \textbf{13.4 $\pm$ 0.2} \\
 & N-Beats \cite{oreshkin2019n} DILATE & 25.8 $\pm$ 0.9  & \textbf{15.5 $\pm$ 0.2} & \textbf{13.3 $\pm$ 0.3} & \textbf{4.4 $\pm$ 0.2} & 3.1 $\pm$ 0.5  & \textbf{13.2 $\pm$ 0.2}  \\
  \cdashline{2-8}
   & Informer \cite{zhou2020informer} MSE & \textbf{38.1 $\pm$ 2.1} & 18.9 $\pm$ 0.6 & 13.2 $\pm$ 0.2 & 6.5 $\pm$ 0.3 & 2.1 $\pm$ 0.2 & 16.4 $\pm$ 0.5  \\
 & Informer \cite{zhou2020informer} DILATE & \textbf{37.8 $\pm$ 0.8} & \textbf{18.5 $\pm$ 0.3} & \textbf{12.9 $\pm$ 0.2} & \textbf{5.7 $\pm$ 0.2} & \textbf{1.9 $\pm$ 0.1}  & \textbf{15.9 $\pm$ 0.3} \\ 
   \midrule 
  ETTH1 (96) & N-Beats \cite{oreshkin2019n} MSE & 32.5 $\pm$ 1.4 & 3.9 $\pm$ 0.2 & 13.3 $\pm$ 2.0 & 21.6 $\pm$ 4.3 & \textbf{5.7 $\pm$ 0.7} & 7.4 $\pm$ 1.0  \\
   & N-Beats \cite{oreshkin2019n} DILATE &  \textbf{26.0 $\pm$ 2.8} & \textbf{2.9 $\pm$ 0.1} & \textbf{4.6 $\pm$ 0.6} & \textbf{11.4 $\pm$ 1.7} & \textbf{6.4 $\pm$ 1.0} & \textbf{4.6 $\pm$ 0.4}  \\
     \cdashline{2-8}
    & Informer \cite{zhou2020informer} MSE & \textbf{28.2 $\pm$ 2.6} & 4.3 $\pm$  0.3 & 5.8 $\pm$ 0.1 & 21.6 $\pm$ 3.3 & \textbf{6.6 $\pm$ 1.9} & 7.8 $\pm$ 0.9    \\
 & Informer \cite{zhou2020informer} DILATE & 32.5 $\pm$ 3.8 & \textbf{3.2 $\pm$ 0.3} & \textbf{4.5 $\pm$ 0.3} & \textbf{19.1 $\pm$ 1.9} & \textbf{6.4 $\pm$ 1.0} & \textbf{6.4 $\pm$ 0.6}    \\
 \bottomrule
    \end{tabular}
    \label{tab:dilate_sota}
\end{table*}

\subsection{Datasets}
\label{sec:datasets}

\begin{figure*}
    \centering
    \includegraphics[width=14.5cm]{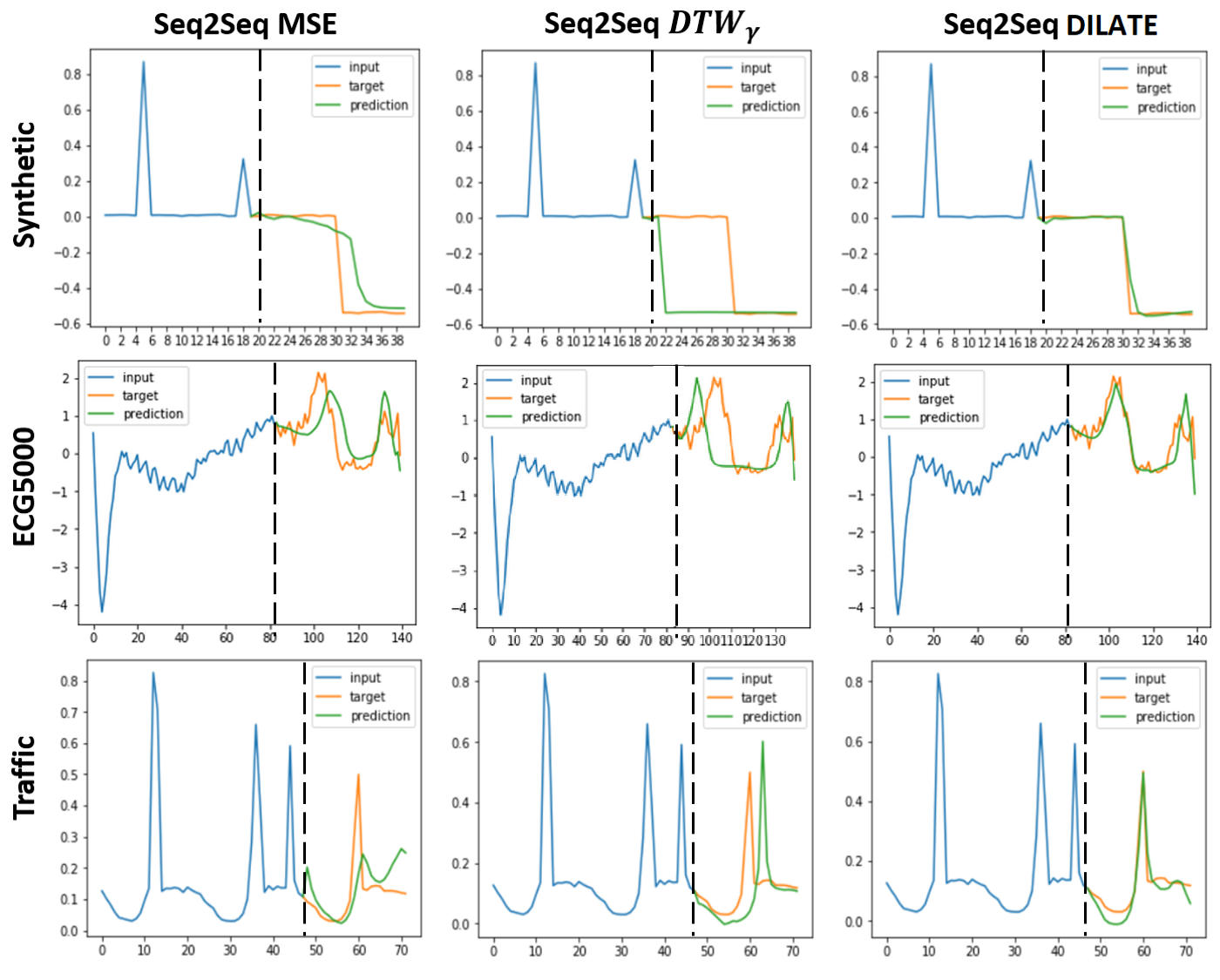}
    \caption{\textbf{Qualitative prediction results with the DILATE loss.} For each dataset, the MSE training loss leads to non-sharp predictions, whereas the soft-DTW loss can predict sharp variations but has no control over their temporal localization. In contrast, the DILATE loss produces sharp predictions with accurate temporal localization.}
    \label{fig:dilate_visu}
\end{figure*}

We carry out experiments on 6 synthetic and real-world datasets from various domains to illustrate the broad applicability of our methods. For each dataset, the task is to predict the $\tau$-steps ahead future trajectory (or multiple trajectories in the probabilistic case) given a $T$-steps context window:

\begin{itemize}
    \item \textbf{Synthetic-det} ($T=20, \tau=20$): deterministic dataset consisting in predicting sudden changes (step functions) based on an input signal composed of two peaks. This controlled setup was designed to measure precisely the shape and time errors of predictions (see Figure \ref{fig:dilate_visu}).
    \item \textbf{Synthetic-prob} ($T=20, \tau=20$): probabilistic version of the previous dataset where for each input series, there exists 10 admissible futures obtained by varying the amplitude and the timing of the step (see Figure \ref{fig-intro} (b)).
    \item \textbf{ECG5000}  ($T=84, \tau=56$): dataset composed of 5000 electrocardiograms (ECG) with sharp spikes whose time intervals are of great importance (see Figure \ref{fig:dilate_visu}).
    \item \textbf{Traffic} ($T=168, \tau=24$) dataset consisting in hourly occupancy rates in California (see Figure \ref{fig:dilate_visu}).
    \item  \textbf{Electricity} ($T=168, \tau=24$): consisting in hourly electricity consumption measurements (kWh) from 370 customers. 
    \item \textbf{ETTh1} \cite{zhou2020informer} ($T=96, \tau=96$):  dataset of hourly Electricity Transformer Temperature measurements, which is an important indicator for electricity grids. This dataset enables to assess the generalization of our approach on much longer term predictions.
\end{itemize}
We give more details for each dataset in Appendix C.

\subsection{Deterministic forecasting with the DILATE loss}

To evaluate the benefits of our proposed DILATE training loss, we compare it against the widely used Euclidean (MSE) loss, and the smooth DTW introduced in~\cite{cuturi2017soft,mensch2018differentiable}. We use the following multi-step prediction metrics: MSE, DTW (shape), TDI (temporal). To consolidate the evaluation, we also consider two additional (non differentiable) metrics for assessing shape and time.  For shape, we compute the ramp score \cite{vallance2017towards}. For time, we compute the Hausdorff distance between a set of detected change points in the target signal $\mathcal{T}^*$ and in the predicted signal 
$\hat{\mathcal{T}}$:
\begin{equation}
\text{Hausdorff}(\mathcal{T}^*,\hat{\mathcal{T}}) :=  \max (  \underset{\hat{t} \in \mathcal{ \hat{T} }}{\max}  \underset{t^* \in \mathcal{ T^* }}{\min} |\hat{t}-t^* |  ,  \underset{t^* \in \mathcal{ T^* }}{\max}   \underset{\hat{t} \in \mathcal{ \hat{T} }}{\min} |\hat{t}-t^* | )
\end{equation}{}
which corresponds to the largest possible distance between a change point and its prediction. Additional details about these external metrics, the architectures and implementation of models used in this section are given in  Appendix D.

\subsubsection{DILATE performances on generic architectures}

To demonstrate the broad applicability of our approach, we first perform multi-step forecasting with two generic neural network architectures: a fully connected network (1 layer of 128 neurons), which does not make
any assumption on data structure, and a more specialized Seq2Seq model with 1 layer of 128 Gated Recurrent Units (GRU). We perform a Student t-test with significance level 0.05 to highlight the best(s) method(s) in each experiment (averaged over 10 runs). Overall results are presented in Table \ref{results1}. \\

\textbf{Comparison to MSE training loss:} DILATE outperforms MSE when evaluated on shape (DTW) in all experiments,  with significant differences on 5/6 experiments. When evaluated on time (TDI), DILATE also performs better in all experiments (significant differences on 3/6 tests). Finally, DILATE is equivalent to MSE when evaluated on MSE on 3/6 experiments.\\

\textbf{Comparison to $\text{DTW}_{\gamma}^{\mathbf{\Delta}}$ training loss:} When evaluated on shape (DTW), DILATE performs similarly to  $\text{DTW}_{\gamma}^{\mathbf{\Delta}}$ (2 significant improvements, 1 significant drop and 3 equivalent performances). For time (TDI) and MSE evaluations, DILATE is significantly better than  $\text{DTW}_{\gamma}^{\mathbf{\Delta}}$ in all experiments, as expected.

We can notice that the ramp score (resp. the Haussdorff distance) provides the same trends than the shape metric DTW (resp. the time metric TDI). It reinforces our conclusions and shows that DILATE indeed improves shape and temporal accuracy beyond the metrics being optimized.

We display a few qualitative examples for Synthetic, ECG5000 and Traffic datasets in Figure \ref{fig:dilate_visu} (other examples are provided in Appendix D.5). We see that MSE training leads to predictions that are non-sharp, making them inadequate in presence of drops or sharp spikes. $\text{DTW}_{\gamma}^{\mathbf{\Delta}}$ leads to very sharp predictions in shape, but with a possibly large temporal misalignment. In contrast, our DILATE loss predicts series that have both a correct shape and precise temporal localization.\\

\subsubsection{DILATE performances with state-of-the-art models}

Beyond generic forecasting architectures, we show that DILATE can also improve the performances of state-of-the-art deep architectures. We experiment here with two recent and competitive models:  N-Beats \cite{oreshkin2019n} and Informer \cite{zhou2020informer}.  Results in Table \ref{tab:dilate_sota} are consistent with those in Table \ref{results1}: models trained with DILATE improve over MSE in shape (in DTW and ramp score for 6/6 experiments) and time (in TDI for 5/6 and Hausdorff for 4/6 experiments) and are equivalent to MSE when evaluated in MSE (equivalent or better for 3/6 experiments). We provide qualitative predictions of N-Beats on Electricity in Appendix D.3.

\subsubsection{DILATE loss analysis \label{sec:dilate-analysis}}

\begin{figure}[H]
    \centering
    \includegraphics[height=4.5cm]{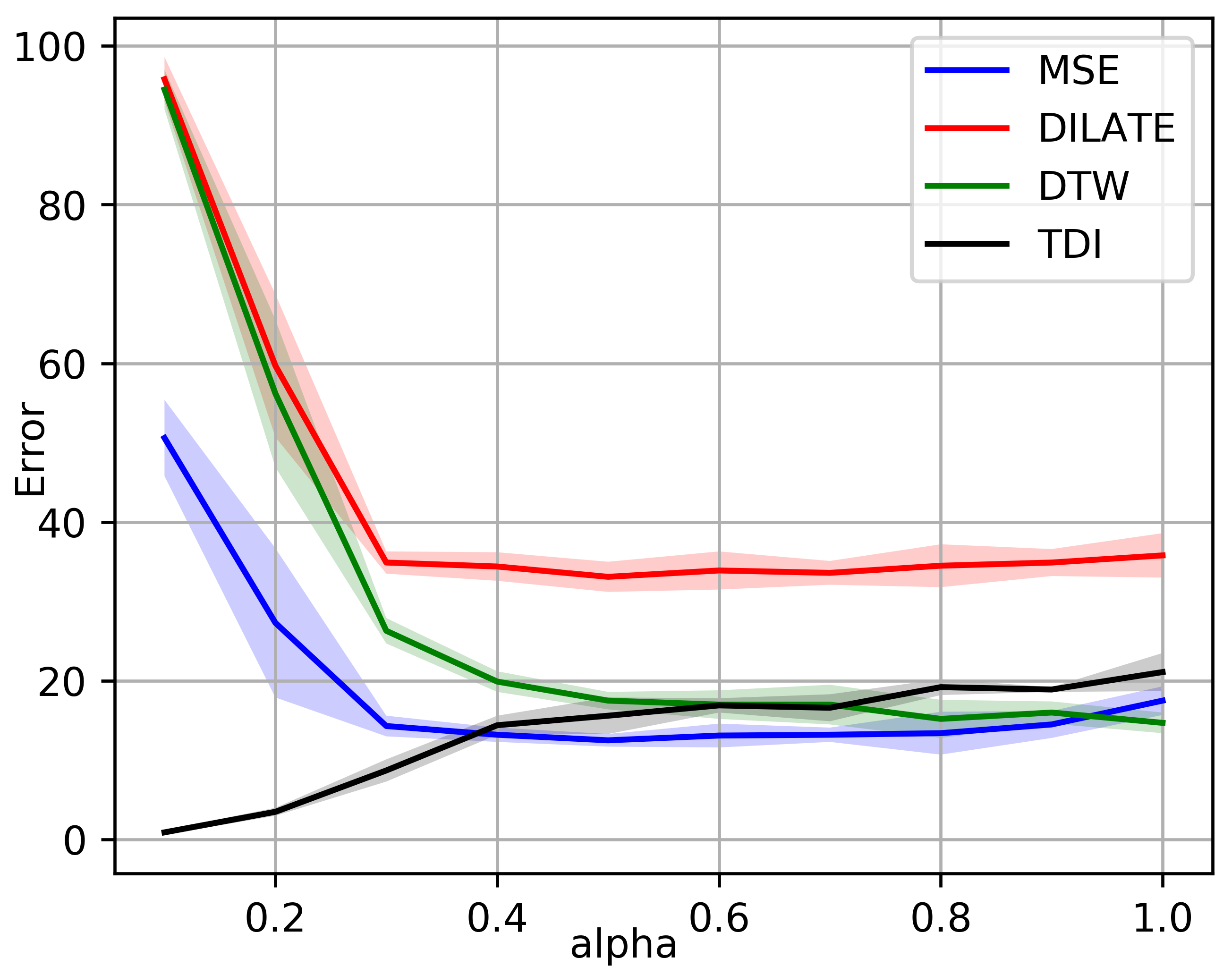}
    \caption{\textbf{Influence of $\alpha$ in the DILATE loss.} The shaded areas represent $\pm$ std computed over 10 runs.}
    \label{fig:influ_alpha-}
\end{figure}

\textbf{Impact of $\alpha$}: We analyze in Figure \ref{fig:influ_alpha-} the influence of the tradeoff parameter $\alpha$ when training a Seq2Seq model on the synthetic-det dataset. When $\alpha=1$, $\mathcal{L}_{\text{DILATE}}$ reduces to $\text{DTW}_{\gamma}^{\mathbf{\Delta}}$, with an accurate shape but a large temporal error. When $\alpha \longrightarrow 0$, we only minimize  $\mathcal{L}_{time}$ without any shape constraint. Both MSE and shape errors explode in this case, illustrating the fact that $\mathcal{L}_{time}$ is only meaningful in conjunction with $\mathcal{L}_{shape}$. Both the MSE and DILATE error curves present a U-shape ; in this case, $\alpha=0.5$ seems an acceptable tradeoff for the Synthetic-det dataset. \\

\begin{figure}[H]
\begin{tabular}{c}
    \includegraphics[height=4.5cm]{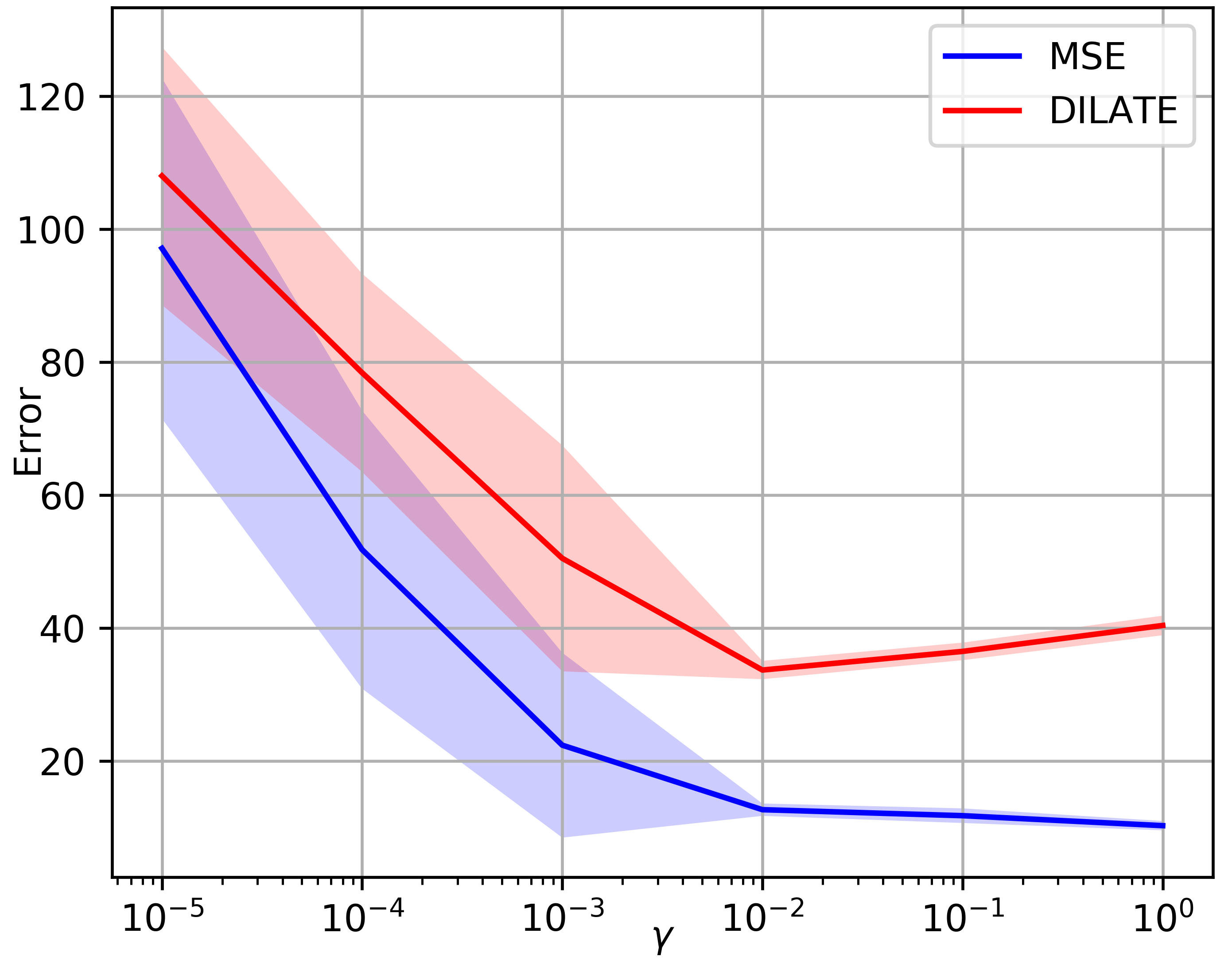}  \\
     \includegraphics[width=8.5cm]{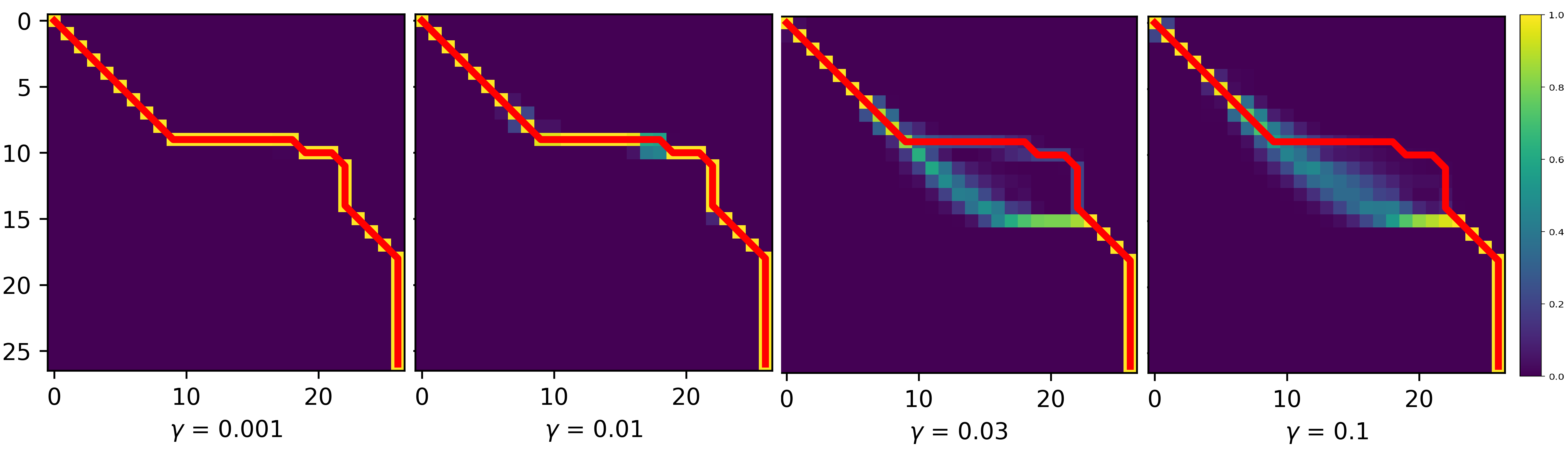}
\end{tabular}
    \caption{\textbf{Influence of $\gamma$ in the DILATE loss.} The shaded areas represent $\pm $ std computed over 10 runs.}
    \label{fig:influ_gamma}
\end{figure}

\textbf{Influence of $\gamma$}: We analyse the influence of the $\text{DTW}_{\gamma}^{\mathbf{\Delta}}$ smoothing parameter $\gamma$ in Figure \ref{fig:influ_gamma}. We represent (bottom of the figure) the assignment probabilities of the $\text{DTW}_{\gamma}^{\mathbf{\Delta}}$ path between two given test time series, the true DTW path being depicted in red. When $\gamma$ increases, the  $\text{DTW}_{\gamma}^{\mathbf{\Delta}}$ path is more uncertain and becomes multimodal. When $\gamma \longrightarrow 0$, the soft DTW converges toward the true DTW. However, we see (top of the figure) that for small $\gamma$ values, optimizing $\text{DTW}_{\gamma}^{\mathbf{\Delta}}$  becomes more difficult, resulting in higher test error and higher variance (on the synthetic-det dataset). We fixed $\gamma=10^{-2}$ in all our experiments, which yields a good tradeoff between an accurate soft optimal path and a low test error.

\begin{table*}[t]
        \caption{\textbf{STRIPE++ forecasting results on the synthetic dataset with multiple futures}, averaged over 5 runs (mean $\pm$ std). Best equivalent methods (Student t-test) shown in bold. Metrics are scaled (MSE $\times$ 1000, DILATE $\times 100$, CRPS $\times$ 1000). \label{table1}  }   
    \begin{adjustbox}{max width=\linewidth}        
    \begin{tabular}{ccccccccccc}
       \toprule 
   \multicolumn{1}{c}{}  & \multicolumn{3}{c}{$\text{H}_{quality}(\cdot) \; (\downarrow)$} & \multicolumn{3}{c}{$\text{H}_{diversity}(\cdot) \; (\downarrow)$} & \multicolumn{3}{c}{$ F1 \text{ score} \; (\downarrow)$}   &  \multicolumn{1}{c}{CRPS ($\downarrow$)} \\ 
       \cmidrule(lr){2-4}   \cmidrule(lr){5-7}   \cmidrule(lr){8-10} 
 Methods   & DTW & TDI   & DILATE  & DTW & TDI  & DILATE & DTW & TDI & DILATE \\ 
        \midrule
        DeepAR \cite{salinas2017deepar} & 42.9 $\pm$ 6.6 & 16.6 $\pm$ 7.6 & 33.5 $\pm$ 6.0 & 23.9 $\pm$ 3.5 & \textbf{12.8 $\pm$ 2.5} & 22.7 $\pm$ 2.2 & 30.7 & 14.5 & 27.1  & 62.4 $\pm$ 9.9   \\
  cVAE DILATE   & \textbf{11.7 $\pm$ 1.5}  &  9.4 $\pm$ 2.2   & \textbf{14.2 $\pm$ 1.5} & 18.8 $\pm$ 1.3 & 48.6 $\pm$ 2.2   &   33.9 $\pm$ 3.9 &  14.4  &  15.7  & 20.0  & 62.2 $\pm$ 4.2  \\
   variety loss \cite{thiede2019analyzing} DILATE  & 15.6 $\pm$ 3.4  & 10.2 $\pm$ 1.1  & 16.8 $\pm$ 0.9 & 22.7 $\pm$ 4.1  & 37.7 $\pm$ 4.9   & 30.8 $\pm$ 1.0 & 18.5  & 16.1 & 21.7  & 62.6 $\pm$ 3.0  \\
    entropy reg. \cite{dieng2019prescribed} DILATE  & 13.8 $\pm$ 3.1  & 8.8 $\pm$ 2.2  & \textbf{15.0 $\pm$ 1.6} & 20.4 $\pm$ 2.8  & 42.0 $\pm$ 7.8  & 32.6 $\pm$ 2.3 & 16.5  & 14.5 & 20.5 & 62.4 $\pm$ 3.9   \\
  Diverse DPP \cite{yuan2019diverse} DILATE  &  \textbf{12.9 $\pm$ 1.2}  & 9.8 $\pm$ 2.1  & 15.1 $\pm$ 1.5 & 18.6 $\pm$ 1.6 &  42.8 $\pm$ 10.1 & 31.3 $\pm$ 5.7 & 15.2 & 15.9 & 20.4 & 60.7 $\pm$ 1.6 \\
 GDPP  \cite{elfeki2018gdpp} DILATE   &  14.8 $\pm$ 2.9  & 11.7 $\pm$ 8.4  &    \textbf{14.4 $\pm$ 2.1} & 20.8 $\pm$ 2.4  & 25.2 $\pm$ 7.2  &   23.9 $\pm$ 4.5  & 17.3  & 15.9 & 17.9 & 63.4 $\pm$ 6.4  \\
 STRIPE \cite{leguen20stripe}   &  16.8 $\pm$ 0.5  & \textbf{6.7 $\pm$ 0.4}  & 15.4 $\pm$ 0.5 & 16.1 $\pm$ 1.1  & \textbf{13.2 $\pm$ 1.7}  & \textbf{17.7 $\pm$ 0.6} & 16.4  & \textbf{8.8}  & 16.5  &  60.5 $\pm$ 0.4 \\ 
 \rowcolor{Gray}
STRIPE++  & 13.5 $\pm$ 0.5 & 9.2 $\pm$ 0.5 & \textbf{15.0 $\pm$ 0.3} & \textbf{12.9 $\pm$ 0.3} & 16.3 $\pm$ 1.2 & \textbf{17.9 $\pm$ 0.6} & \textbf{13.2} & 11.7 & \textbf{16.3} & \textbf{48.6 $\pm$ 0.6}   \\
\bottomrule
    \end{tabular}
    \end{adjustbox}
    \label{tab:stripe}
\end{table*}{}

\subsection{Probabilistic forecasting with STRIPE++\label{sec:prob-for}}

We firstly assess the ability of STRIPE++ to capture the full predictive distribution of future trajectories. To do so, we need for evaluation the ground truth set of admissible futures for a given input ; we use here the synthetic-prob dataset (described in section \ref{sec:datasets}) constructed for this purpose. Secondly, on a more realistic setting where we only know one future for each input, we evaluate STRIPE++ on the Traffic and Electricity datasets with the best (resp. the mean) sample metrics as a proxy for diversity (resp. quality). We describe the implementation details and neural network architectures (encoder, decoder, posterior net and STRIPE++ proposal network) in Appendix E.2.

\subsubsection{Full predictive distribution evaluation on synthetic-prob}
\label{sec:stripe-synth}


\textbf{Metrics:} To assess the discrepancy between the predicted and true distributions of futures trajectories, we define two measures $\text{H}_{quality}(\ell)$ and $\text{H}_{diversity}(\ell)$ ($\ell = $ DTW, TDI or DILATE in our experiments):
\begin{align}
        \text{H}_{quality}(\ell) &:= \mathbb{E}_{\x \in \mathcal{D}_{test}} \mathbb{E}_{\hat{\y}}  \left[  \underset{\y \in F(\x)}{\inf} \: \ell(\hat{\y},\y) \right]   \\
         \text{H}_{diversity}(\ell) &:= \mathbb{E}_{\x \in \mathcal{D}_{test}} \mathbb{E}_{\y \in F(\x)}  \left[  \underset{\hat{\y}}{\inf} \: \ell(\hat{\y},\y) \right] \\
         F1 \text{ score} &= \frac{2 ~ \text{H}_{quality}(\ell)  \cdot  \text{H}_{diversity}(\ell) }{ \text{H}_{quality}(\ell) +  \text{H}_{diversity}(\ell)} \label{eq:F1score}
\end{align}
$\text{H}_{quality}$ penalizes forecasts $\hat{\y}$ that are far away from a ground truth future of $\x$ denoted $\y \in F(\x)$ (similarly to the \textit{precision} concept in pattern recognition) whereas $\text{H}_{diversity}$ penalizes when a true future is not covered by a forecast (similarly to \textit{recall}). As a tradeoff balancing quality and diversity, we compute the F1 score in Eq. \ref{eq:F1score}. In addition, we also use the continuous ranked probability score (CRPS)  which is a standard \textit{proper scoring rule} \cite{gneiting2007probabilistic} for assessing probabilistic forecasts \cite{gasthaus2019probabilistic}. Intuitively, the CRPS is the pinball loss integrated over all quantile levels. A key property is that the CRPS attains its minimum when the predicted future distribution equals the true future distribution, making this metric particularly adapted to our context.\\

\begin{table*}
\caption{\textbf{Probabilistic forecasting results on the Traffic and Electricity datasets}, averaged over 5 runs (mean $\pm$ std). Metrics are scaled for readability. Best equivalent method(s) (Student t-test) shown in bold.} 
   \label{tab:stripe_sota}    
 \centering              
\setlength{\tabcolsep}{6.8pt}
    \begin{adjustbox}{max width=\linewidth}
    \begin{tabular}{ccccc|cccc}
    \toprule
    \multicolumn{1}{c}{} & \multicolumn{4}{c|}{Traffic} & \multicolumn{4}{c}{Electricity}   \\
    \multicolumn{1}{c}{} & \multicolumn{2}{c}{MSE} &  \multicolumn{2}{c|}{DILATE}   & \multicolumn{2}{c}{MSE} &  \multicolumn{2}{c}{DILATE}  \\ 
         \cmidrule(lr){2-3}   \cmidrule(lr){4-5}      \cmidrule(lr){6-7}   \cmidrule(lr){8-9} 
 Method   & mean   & best & mean & best   & mean & best & mean  & best   \\ 
 \midrule
 Nbeats \cite{oreshkin2019n} MSE  &  - & 7.8 $\pm$ 0.3 & - & 22.1 $\pm$ 0.8  & - & 24.8 $\pm$ 0.4 & - & 20.2 $\pm$ 0.3    \\
 Nbeats \cite{oreshkin2019n} DILATE  & - & 17.1 $\pm$ 0.8 & - & 17.8 $\pm$ 0.3  & - & 25.8 $\pm$ 0.9 & - & 19.9 $\pm$ 0.5 \\
  \midrule
   Deep AR \cite{salinas2017deepar} & 15.1 $\pm$ 1.7 & \textbf{6.6 $\pm$ 0.7} & 30.3 $\pm$ 1.9 & 16.9 $\pm$ 0.6  &   67.6 $\pm$ 5.1  & 25.6 $\pm$ 0.4 & 59.8 $\pm$ 5.2  & 17.2 $\pm$ 0.3    \\
   cVAE DILATE & \textbf{10.0 $\pm$ 1.7}  & 8.8 $\pm$ 1.6  & \textbf{19.1 $\pm$ 1.2} & 17.0 $\pm$ 1.1   &  \textbf{28.9 $\pm$ 0.8}    &  27.8 $\pm$ 0.8 & 24.6 $\pm$ 1.4   & 22.4 $\pm$ 1.3   \\
   Variety loss \cite{thiede2019analyzing} & \textbf{9.8 $\pm$ 0.8} & 7.9 $\pm$ 0.8  & \textbf{18.9 $\pm$ 1.4}  & 15.9 $\pm$ 1.2  & 29.4 $\pm$ 1.0  & 27.7 $\pm$ 1.0  & 24.7 $\pm$ 1.1  & 21.6 $\pm$ 1.0    \\
   Entropy regul. \cite{dieng2019prescribed} & 11.4 $\pm$ 1.3   & 10.3 $\pm$ 1.4  & \textbf{19.1 $\pm$ 1.4}  & 16.8 $\pm$ 1.3  & 34.4 $\pm$ 4.1  &  32.9 $\pm$ 3.8  & 29.8 $\pm$ 3.6  & 25.6 $\pm$ 3.1  \\
   Diverse DPP \cite{yuan2019diverse}  & 11.2 $\pm$ 1.8  & 6.9 $\pm$ 1.0   & 20.5 $\pm$ 1.0  & 14.7 $\pm$ 1.0   &  31.5 $\pm$ 0.8  & 25.8 $\pm$ 1.3  & 26.6 $\pm$ 1.0  & 19.4 $\pm$ 1.0    \\ 
  STRIPE \cite{leguen20stripe} & \textbf{10.1 $\pm$ 0.4}   &  \textbf{6.5 $\pm$ 0.2}  & \textbf{19.2 $\pm$ 0.8}  & \textbf{14.2 $\pm$ 0.2}   & 29.7 $\pm$ 0.3  & \textbf{23.4 $\pm$ 0.2}  & 24.4  $\pm$ 0.3  & \textbf{16.9 $\pm$ 0.2}   \\
 \rowcolor{Gray} 
  STRIPE++ & \textbf{10.0 $\pm$ 0.2} & \textbf{6.7 $\pm$ 0.3} & \textbf{19.0 $\pm$ 0.2} & \textbf{14.1 $\pm$ 0.3}  & \textbf{29.5 $\pm$ 0.3} & \textbf{23.6 $\pm$ 0.4} & \textbf{24.1 $\pm$ 0.2}  & 17.3 $\pm$ 0.4  \\
  \bottomrule
    \end{tabular}
    \end{adjustbox}
\end{table*}{}

\textbf{Forecasting results:} We compare in Table \ref{tab:stripe} our method to 4 recent competing diversification mechanisms (variety loss \cite{thiede2019analyzing}, entropy regularisation \cite{dieng2019prescribed}, diverse DPP \cite{yuan2019diverse} and GDPP \cite{elfeki2018gdpp}) based on a conditional variational autoencoder (cVAE) backbone trained with DILATE. We observe that STRIPE and STRIPE++ obtain the global best performances by improving diversity by a large amount ($\text{H}_{diversity}(\text{DILATE)}$=17.6) compared to the backbone cVAE DILATE ($\text{H}_{diversity}(\text{DILATE)}$=33.9) and to other diversification schemes (the best competitor GDPP \cite{elfeki2018gdpp} attains $\text{H}_{diversity}(\text{DILATE)}$=23.9).
This highlights the relevance of the structured shape and time diversity. We can also notice that, in contrast to competing diversification schemes that improve diversity at the cost of a loss in quality, STRIPE++ maintains high quality predictions. STRIPE++ is only beaten in $\text{H}_{quality}(\text{DILATE)}$ by GDPP \cite{elfeki2018gdpp}, but this method is significantly worse than STRIPE++ in diversity, and GDPP requires full future distribution supervision, which it not applicable in real datasets (see section \ref{sec:stripe_real_datasets}). All in all, the F1 scores summarize the quality \vs diversity tradeoffs, and STRIPE++ gets the best F1 DILATE score. Moreover, STRIPE++ outperforms all other methods with the CRPS metric, indicating that the predicted future trajectory distribution is closer to the ground truth one.\\

\textbf{Discussion on quality regularization:} As discussed above, the key difference between STRIPE and STRIPE++ consists in an additional quality constraint during the diversification stage. We observe that STRIPE++ indeed produces globally better quality forecasts ($\text{H}_{quality}(\text{DILATE)}$= 15.0 for STRIPE++ \vs 15.5 for STRIPE) while maintaining the level of diversity. It leads to a better DILATE F1 score and a better CRPS for STRIPE++.

\subsubsection{Ablation study}

To analyze the respective roles of the quality and diversity losses, we perform an ablation study on the synthetic-prob dataset with the cVAE backbone trained with the quality loss DILATE and different DPP diversity losses. For a finer analysis, we report in Table \ref{tab:ablation} the shape (DTW, computed with Tslearn \cite{tavenard2020tslearn}) and time (TDI) components of the DILATE loss \cite{leguen19dilate}.

Results presented in Table \ref{tab:ablation} first reveal the crucial importance to define different criteria for quality and diversity. With the same loss for quality and diversity (as this is the case in \cite{yuan2019diverse}), we observe here that the DILATE DPP kernel $\exp(-\text{DILATE})$ does not bring a significant diversity gain compared to the cVAE DILATE baseline (without diversity loss). By choosing the MSE kernel instead, we even get a small diversity and quality improvement.
 
In contrast, our introduced shape and time kernels $\mathcal{K}_{shape}$ and  $\mathcal{K}_{time}$ largely improve the diversity in DILATE without deteriorating quality. As expected, each kernel brings its own benefits:  $\mathcal{K}_{shape}$ brings the best improvement in the shape metric DTW ($\text{H}_{diversity}(\text{DTW)}=$ 16.4 \vs 18.8) and $\mathcal{K}_{time}$ the best improvement in the time metric TDI ($\text{H}_{diversity}(\text{TDI)}=$ 15.1 \vs 48.6). STRIPE++ gathers the benefits of both criteria and gets the global best results in diversity and equivalent results in quality.

\begin{table}
        \caption{\textbf{Ablation study on the synthetic-prob dataset}. We train a backbone cVAE with the DILATE quality loss and compare different DPP kernels for diversity.}
    \begin{adjustbox}{max width=\linewidth}
    \begin{tabular}{ccccc}
    \toprule
    \multicolumn{1}{c}{} & \multicolumn{1}{c}{$\text{H}_{quality} \; (\downarrow)$} & \multicolumn{3}{c}{$\text{H}_{diversity}(.) \; (\downarrow)$}   \\ 
       \cmidrule(lr){2-2}   \cmidrule(lr){3-5} 
   diversity &  DILATE & DTW & TDI & DILATE     \\ \hline 
 None  & \textbf{14.2 $\pm$ 1.5} &     18.8 $\pm$ 1.3 & 48.6  $\pm$ 2.2  &  33.9 $\pm$ 3.9  \\
 DILATE &     \textbf{15.1 $ \pm$ 1.9} &  
    18.6 $\pm$ 1.6 &   42.8 $\pm$ 10.2 &  31.3 $\pm$ 10.7  \\  
 MSE &   \textbf{15.1 $\pm$ 1.4} &  18.5 $\pm$ 1.3 & 41.9 $\pm$ 8.8 & 30.8 $\pm$ 4.7 \\
  \rowcolor{Gray}
 shape  & \textbf{15.1 $\pm$ 0.6}  & 16.4 $\pm$ 1.5  & 
15.4 $\pm$ 4.2  & 18.9 $\pm$ 1.8  \\
 \rowcolor{Gray}
time  &  \textbf{15.6 $\pm$ 0.6} & 
  17.6 $\pm$ 0.5 &  \textbf{15.1 $\pm$ 3.1}  &  19.4 $\pm$ 1.6  \\
   \rowcolor{Gray}
STRIPE++    & \textbf{15.0 $\pm$ 0.3}   & \textbf{12.9 $\pm$ 0.3} & \textbf{16.3 $\pm$ 1.2} & \textbf{17.9 $\pm$ 0.6}  \\ 
\bottomrule
    \end{tabular}
    \end{adjustbox}
    \label{tab:ablation}    
    \vspace{-0.3cm}
\end{table}{}

\begin{figure*}
\centering
\begin{tabular}{cc}
   \includegraphics[width=8cm]{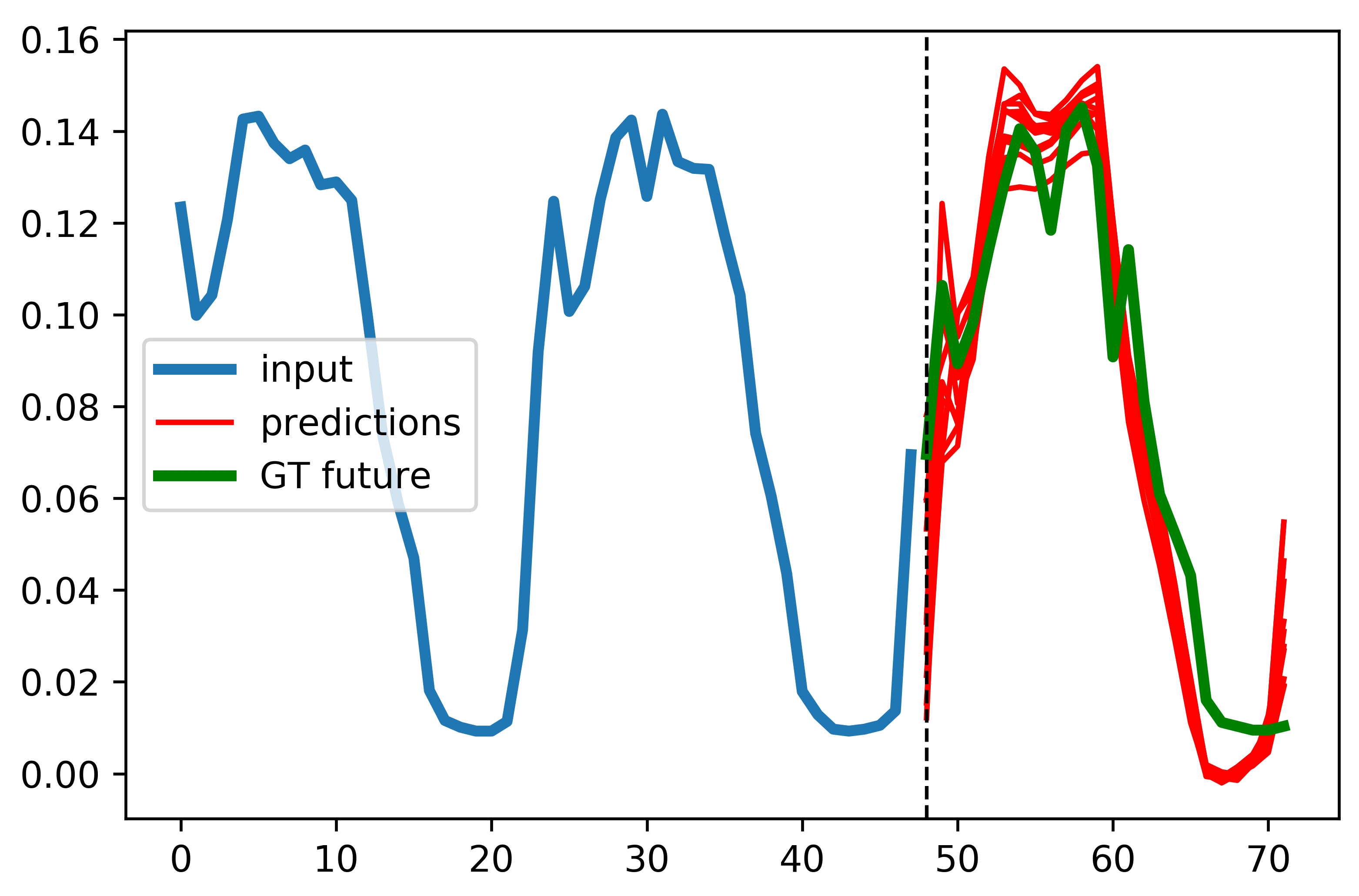}   &    \includegraphics[width=8cm]{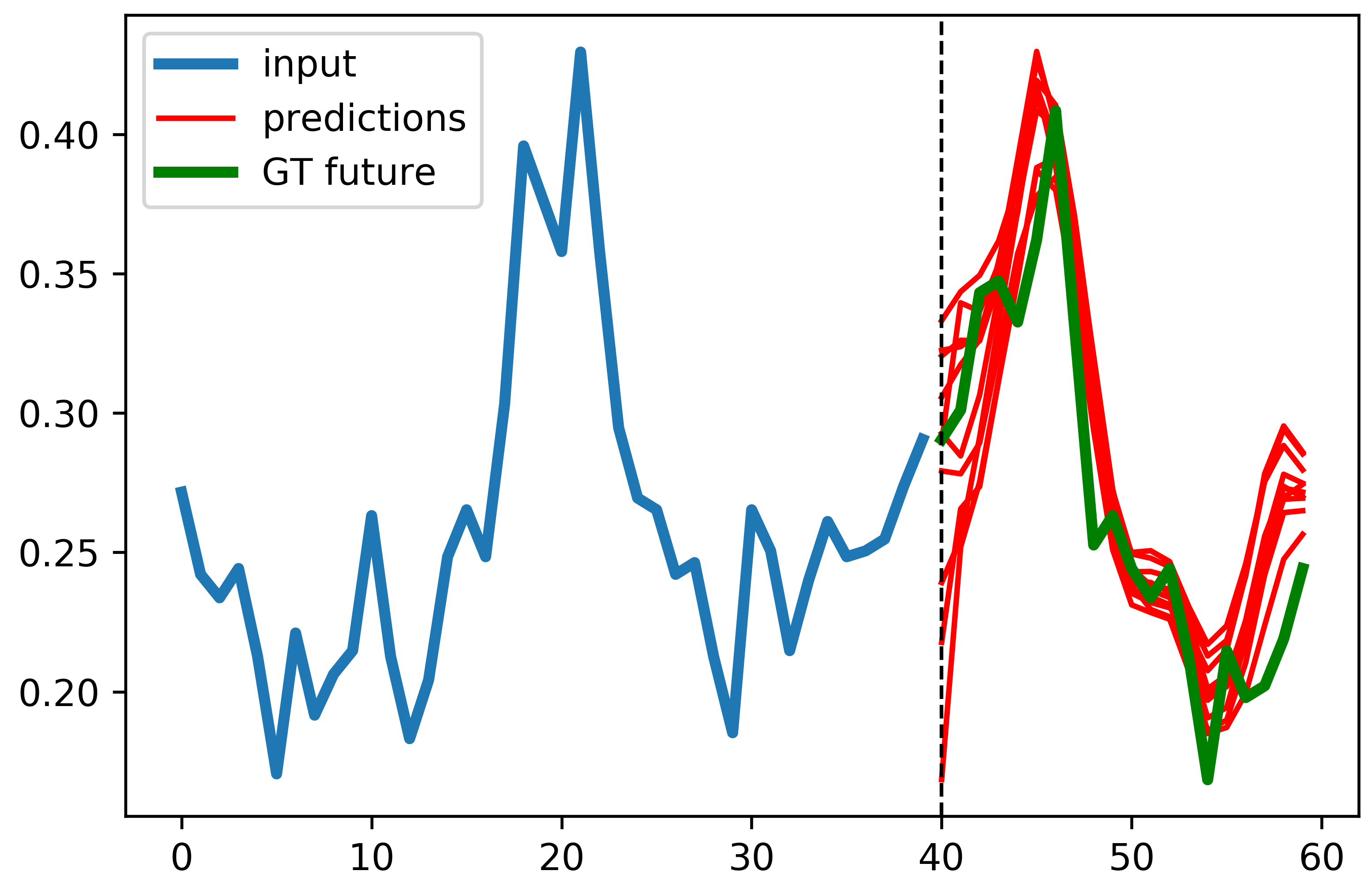} \\
   (a) Traffic & (c) Electricity
\end{tabular}
    \caption{STRIPE++ qualitative predictions on real-world datasets Traffic (a) and Electricity (b).}
    \label{fig:stripe_visus}
\end{figure*}

\subsubsection{State-of-the-art comparison on real-world datasets}
\label{sec:stripe_real_datasets}

We evaluate here the performances of STRIPE++ on two challenging real-world datasets Traffic and Electricity commonly used as benchmarks in the time series forecasting literature \cite{yu2016temporal,salinas2017deepar,lai2018modeling,rangapuram2018deep,leguen19dilate,sen2019think}. Contrary to the synthetic-prob dataset, we only dispose of one future trajectory sample $\y^{*}_{T+1:T+\tau}$ for each input series $\x_{1:T}$. In this case, the metric $\text{H}_{quality}$ (resp. $\text{H}_{diversity}$) defined in section \ref{sec:stripe-synth} reduces to the mean sample (resp. best sample), which are common for evaluating stochastic forecasting models \cite{babaeizadeh2017stochastic,franceschi2020stochastic}. 

Results in Table \ref{tab:stripe_sota} reveal that STRIPE and STRIPE++ outperform all other baselines in the best sample (evaluated in MSE or DILATE). Our method even outperforms in the best sample the state-of-the-art N-Beats algorithm \cite{oreshkin2019n} (either trained with MSE or DILATE), which is dedicated to producing high quality deterministic forecasts. In terms of quality (evaluation with the mean sample), STRIPE++ improves over STRIPE (as expected) and gets the best (or equivalently best) results in all cases. This contrasts to competing diversification methods, \eg Diverse DPP \cite{yuan2019diverse}, that deteriorate quality to improve diversity. Finally we notice that STRIPE++ is consistently better in diversity and quality than the state-of-the art probabilistic deep AR method \cite{salinas2017deepar}.

We display a few qualitative forecasting examples of STRIPE++ on Figure \ref{fig:stripe_visus}.
~We observe that STRIPE++ predictions are both sharp and accurate: both the shape diversity (amplitude of the peaks) and temporal diversity match the ground truth future.

\subsubsection{STRIPE++ analysis: quality-diversity cooperation}

\begin{figure}[H]
    \centering
    \includegraphics[width=6.5cm]{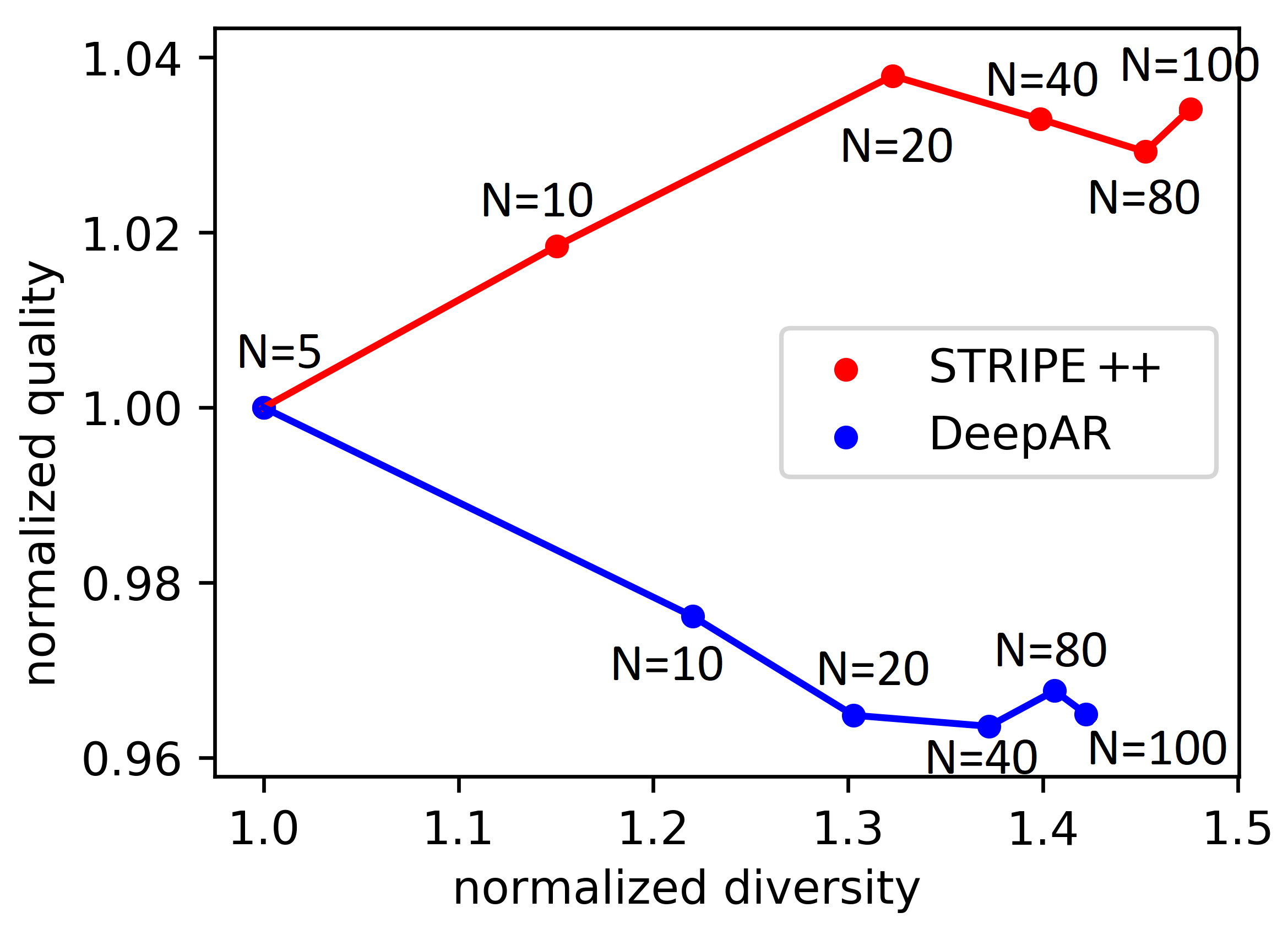}
\caption{Influence of the number $N$ of trajectories on quality (higher is better) and diversity for the \texttt{synthetic-prob} dataset.}    
    \label{fig:stripe_N}
\end{figure}

We analyze here the quality-diversity tradeoff with respect to the  number $N$ of sampled future trajectories. In Figure \ref{fig:stripe_N} we represent the evolution of performances when $N$ increases from 5 to 100 on the synthetic-prob dataset. As expected, the normalized DILATE diversity  $\text{H}_{diversity}(5)/\text{H}_{diversity}(N)$ (higher is better) increases with $N$ for both STRIPE++ and deepAR models \cite{salinas2017deepar}. However we remark that STRIPE++ does not deteriorate normalized quality (which even increases slightly), in contrast to deepAR which does not have control over the targeted diversity. This again confirms the relevance of our approach that effectively combines an adequate quality loss function and a structured diversity mechanism. We provide an additional cooperation analysis in Appendix E.3 highlighting the importance to decouple the criteria used for quality and diversity.

\section{Conclusion}

In this work, we tackle the multi-step deep time series forecasting problem, in the challenging context of non-stationary series that can present sharp variations. In contrast to the majority of existing methods that train models with the surrogate MSE or variants, we propose to leverage shape and temporal criteria at training time. We introduce differentiable similarities and dissimilarities for characterizing shape accuracy  and temporal localization error. We provide two implementations for time series forecasting: the DILATE loss function for deterministic forecasting that ensures both sharp predictions with accurate temporal localization, and the STRIPE++ model for probabilistic forecasting with shape and temporal diversity. We validate our claims with extensive experiments on synthetic and real-world datasets.

An interesting future perspective would be to incorporate seasonality and extrinsic prior knowledge (such as special events) \cite{laptev2017time} to better model the non-stationary abrupt changes and measure their impact on diversity and model confidence. Other appealing directions include diversity-promoting forecasting for exploration in reinforcement learning \cite{pathak2017curiosity,eysenbach2018diversity,leurent2020robust}, and extension of shape/temporal criteria and structured diversity to spatio-temporal or video prediction tasks \cite{xingjian2015convolutional,franceschi2020stochastic,leguen-phydnet,leguen-aphynity}.


%





\bibliographystyle{IEEEtran}
\bibliography{refs.bib}

\begin{IEEEbiography}[{\includegraphics[width=1in,height=1.25in,clip,keepaspectratio]{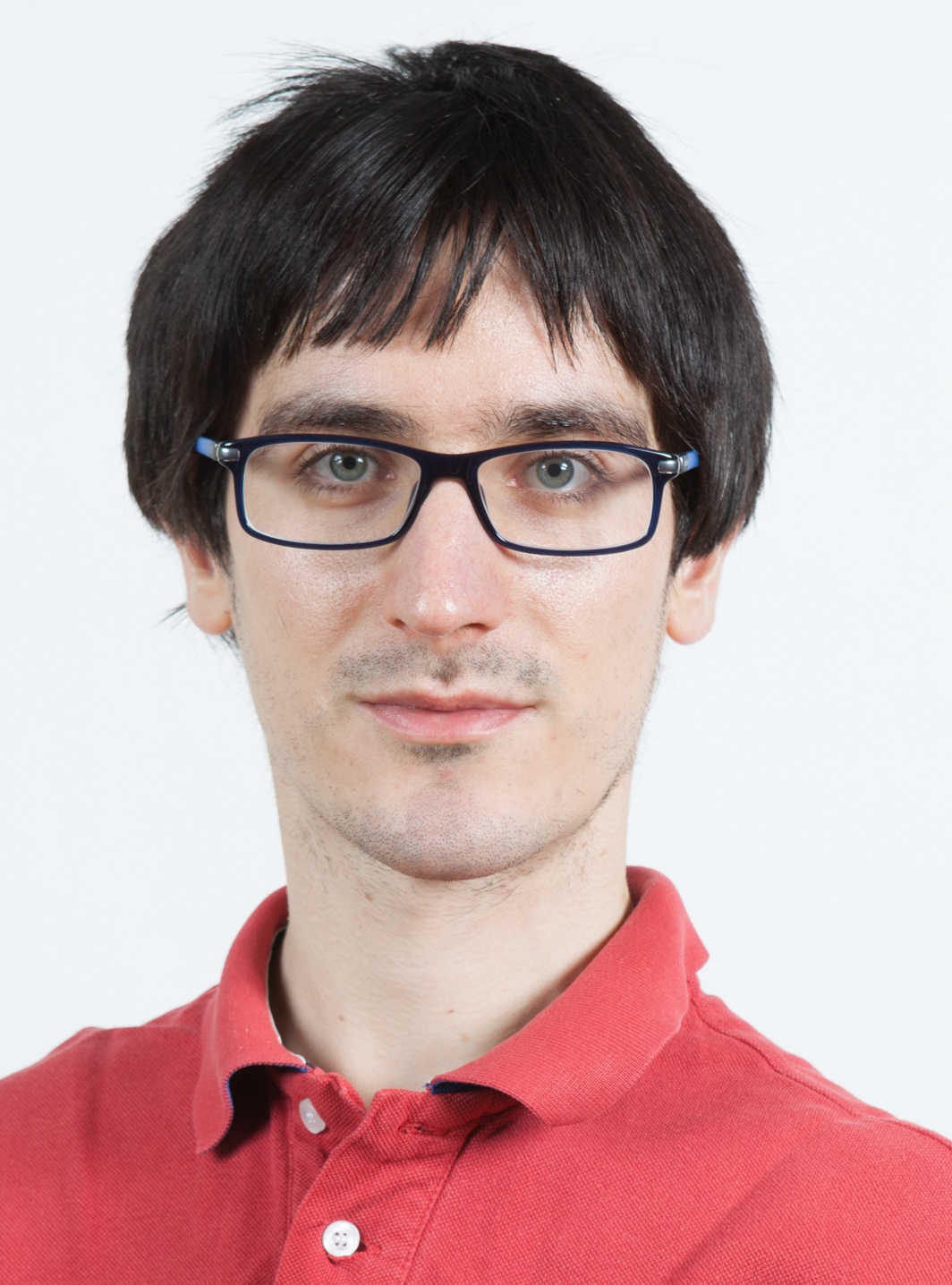}}]{Vincent Le Guen}
is a Ph.D. student in deep learning and computer vision at  Conservatoire National des Arts et
Métiers (CNAM Paris) and EDF research
lab (France). He received a M.Eng. degree from Télécom Paris and a M.Sc degree in applied mathematics from Ecole Normale Supérieure Paris Saclay in 2013. His research interests include deep learning for spatio-temporal phenomena forecasting and physics-inspired machine learning.
\end{IEEEbiography}

\begin{IEEEbiography}[{\includegraphics[width=1in,height=1.25in,clip,keepaspectratio]{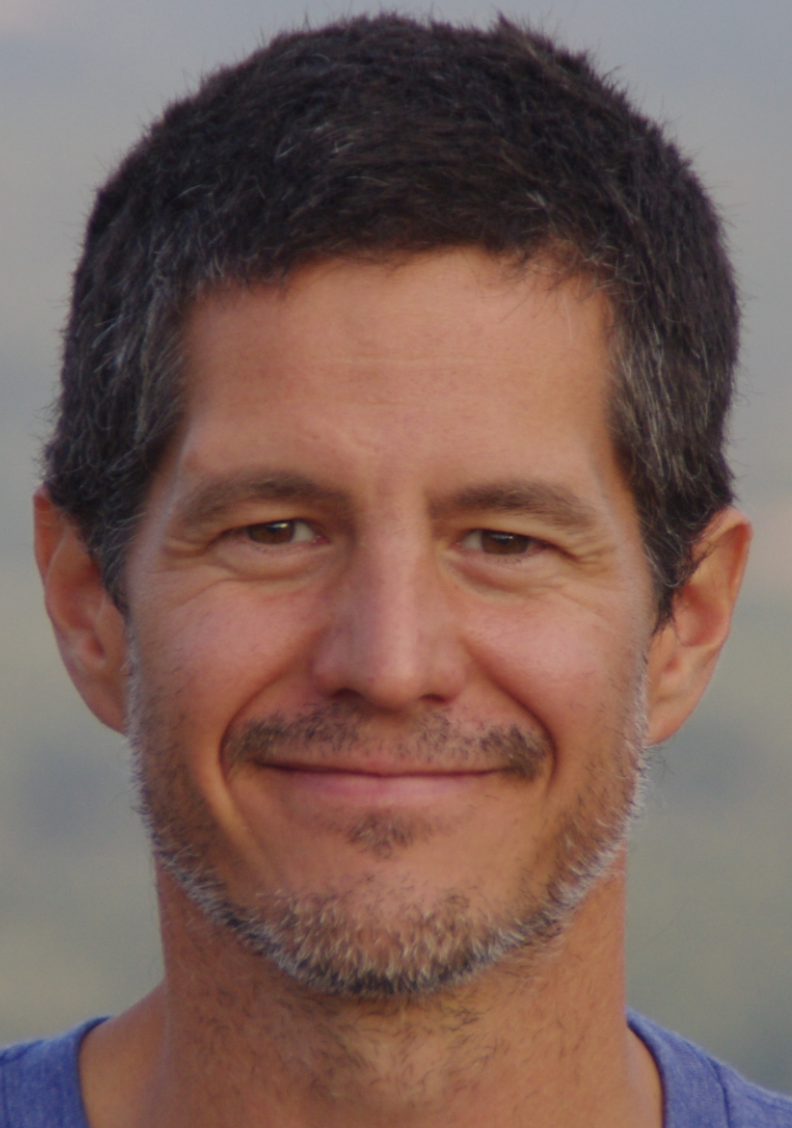}}]{Nicolas Thome}
is a full professor at Conservatoire
Nationnal des Arts et Métiers (CNAM Paris).
His research interests include machine learning
and deep learning for understanding low-level
signals, e.g.. vision, time series, acoustics, etc.
He also explores solutions for combining low-level
and more higher-level data for multi-modal
data processing. His current application domains
are essentially targeted towards healthcare, autonomous
driving and physics. He is involved in
several French, European and international collaborative
research projects on artificial intelligence and deep learning.
\end{IEEEbiography}

\end{document}


%
\title{Deep Time Series Forecasting with Shape and Temporal Criteria \\ Appendix}

\author{Vincent~Le Guen,
        Nicolas~Thome
}

%
%

\markboth{IEEE TRANSACTIONS ON PATTERN ANALYSIS AND MACHINE INTELLIGENCE}%
{Shell \MakeLowercase{\textit{et al.}}: Deep Time Series Forecasting with Shape and Temporal Criteria}
\maketitle

\appendices


\section{Proof that the temporal kernel is PSD}
\label{app:proof-ktime}

The DTW score between two time series $\y \in \mathbb{R}^{d \times n}$ and $\z \in \mathbb{R}^{d \times m}$ can be written $S(\pi) = \sum_{i=1}^{|\pi|} \mathbf{\Delta}(\y_{\pi_1(i)}, \z_{\pi_2(i)})$ where $\pi=(\pi_1,\pi_2)$ is a valid alignment between both series. Equivalently we can write the DTW score $S(\pi) = S(\mathbf{A}) = \left\langle \mathbf{A}, \mathbf{\Delta(\y,\z)} \right\rangle$, where $\mathbf{A} \subset \left \{  0,1 \right \}  ^{n \times m}$ is the warping path in matrix form ($\mathbf{A}_{ij}=1$ if $\y_i$ is associated to $\z_j$ and 0 otherwise).\\

Let $w: \mathcal{A}_{n,m} \longrightarrow \mathbb{R}_+^*$ be a strictly positive weighting function on alignment paths and let's consider the following kernel:
\begin{align}
\mathcal{K}_w(\y,\z) &=    \sum_{\mathbf{A} \in \mathcal{A}_{n,m}} w(\mathbf{A}) ~~ e^{ -  \frac{S(\mathbf{A})}{\gamma} }  \\
&= \sum_{\mathbf{A} \in \mathcal{A}_{n,m}} w(\mathbf{A}) ~~ e^{ -  \frac{\left\langle \mathbf{A} , \mathbf{\Delta(\y,\z)} \right\rangle}{\gamma} } \\
 &=  \sum_{\pi \in \mathcal{A}_{n,m}} w(\pi)  ~~ e^{ -  \frac{ \sum_{j=1}^{|\pi|} \mathbf{\Delta} \left(  \y_{\pi_1(j)} , \z_{\pi_2(j)} \right) }{\gamma} } \\ 
&=  \sum_{\pi \in \mathcal{A}_{n,m}} w(\pi) \prod_{j=1}^{|\pi|} e^{ - \frac{ \mathbf{\Delta} \left(  \y_{\pi_1(j)} , \z_{\pi_2(j)} \right) }{\gamma} } \\
 &=  \sum_{\pi \in \mathcal{A}_{n,m}} w(\pi) \prod_{j=1}^{|\pi|} k(\y_{\pi_1(j)} , \z_{\pi_2(j)}) 
\label{eq:kernel_def}
\end{align}
where we denote $k=e^{-\frac{\mathbf{\Delta}}{\gamma}}$. We prove the following result: \\

\begin{prop}
If $k$ is a PSD kernel such that $\frac{k}{1+k}$ is also PSD, the kernel $\mathcal{K}_w$ defined in Eq. \ref{eq:kernel_def} is also PSD.
\end{prop}

\begin{proof}
The proof is adapted from \cite{cuturi2007kernel}. First, for any time series $\y= (\y_1,\dots,\y_n) \in \mathbb{R}^{d \times n}$ of length $n$ and for any sequence $a \in \mathbb{N}^n$, we introduce the notation:
\begin{equation}
    \y_a = (\underset{a_1 \text{~times}}{\underbrace{\y_1,\dots,\y_1}}, \dots,  \underset{a_n \text{~times}}{\underbrace{\y_n,\dots,\y_n}})
\end{equation}

Let $\chi$ be any PSD kernel defined on $\mathbb{R}^d$ with the following condition $|\chi| < 1$, we introduce the kernel $\kappa$ defined as:
\begin{equation}
    \kappa(\y,\z) = 
    \begin{cases}
    \prod_{i=1}^{|x|} \chi(\y_i, \z_j)  \text{~~if~~} |\y| = |\z| \\
    0 \text{~~~otherwise}
    \end{cases}
\end{equation}

Then, given a strictly positive weighting function $w(a,b) > 0$, the following kernel $\mathcal{K}_w$ defined in Eq. \ref{eq:Kw}  is PSD by construction:
\begin{equation}
    \mathcal{K}_w(\y,\z) = \sum_{a \in \mathbb{N}^n} \sum_{b \in \mathbb{N}^m} w(a,b) ~ \kappa(\y_a, \z_b)
    \label{eq:Kw}
\end{equation}
where we recall that $n=|\y|$ and $m=|\z|$. We denote $\epsilon_a = (\underset{a_1 \text{~times}}{\underbrace{1,\dots,1}}, \dots,  \underset{a_p \text{~times}}{\underbrace{p,\dots,p}})$ for any $a\in \mathbb{N}^p$. We also write for any sequences $u$ and $v$ of common length $p$: $u \otimes v = ((u_1,v_1),\dots,(u_p,v_p))$. With these notations, we can rewrite $\mathcal{K}_w$ as:
\begin{equation}
    \mathcal{K}_w(\y,\z) = \sum_{ \overset{a \in \mathbb{N}^n, b \in \mathbb{N}^m}{\Vert a \Vert = \Vert b \Vert} } w(a,b)  \prod_{i=1}^{\Vert  a \Vert} \chi((\y,\z)_{\epsilon_a \otimes \epsilon_b(i)})
    \label{eq:Kw_ab}
\end{equation}

Notice now for each couple $(a,b)$ there exists a unique alignment path $\pi$ and an integral vector $v$ verifying $\pi_v = \epsilon_a \otimes \epsilon_b$. Conversely, for each couple $(\pi,v)$ there exists a unique pair $(a,b)$ verifying  $\pi_v = \epsilon_a \otimes \epsilon_b$. Therefore the kernel $\mathcal{K}_w$ in Eq. \ref{eq:Kw_ab} can be written equivalently with a parameterization on $(\pi,v)$ for $w$:
\begin{equation}
    \mathcal{K}_w(\y,\z) = \sum_{\pi \in \mathcal{A}_{n,m}} \sum_{v \in \mathbb{N}^{|\pi|}} w(\pi,v) \prod_{j=1}^{|\pi|} \chi((\y,\z)_{\pi_v(j)})
\end{equation}
\label{eq:Kw_piv}
where $\chi_{\pi(j)}$ is a shortcut for $\chi(\y_{\pi_1(j)}, \z_{\pi_2(j)})$.\\

Now we assume that the weighting function $w$ depends only on $\pi$: $w(\pi,v)=w(\pi)$. Then we have:
\begin{align*}
\mathcal{K}_w(\y,\z) &=  \sum_{\pi \in \mathcal{A}_{n,m}} w(\pi) \sum_{v \in \mathbb{N}^{|\pi|}} \prod_{j=1}^{|\pi|} \chi^{v_j}_{\pi(j)} \\
 &= \sum_{\pi \in \mathcal{A}_{n,m}} w(\pi) \prod_{j=1}^{|\pi|} \left( \chi_{\pi(j)} + \chi_{\pi(j)}^2  + \chi_{\pi(j)}^3 + \dots   \right)\\
  &=  \sum_{\pi \in \mathcal{A}_{n,m}} w(\pi) \prod_{j=1}^{|\pi|} \frac{\chi_{\pi(j)}}{1-\chi_{\pi(j)}}
\end{align*}
By setting now $\chi = \frac{k}{1+k}$ which is PSD by hypothesis and verifies $| \chi | <1$ (recall that $k=e^{- \frac{\mathbf{\Delta}}{\gamma}} $), we get:

\begin{align*}
\mathcal{K}_w(\y,\z) &=  \sum_{\pi \in \mathcal{A}_{n,m}} w(\pi) \prod_{j=1}^{|\pi|} k_{\pi(j)} \\
 &=  \sum_{\pi \in \mathcal{A}_{n,m}} w(\pi) \prod_{j=1}^{|\pi|} k(\y_{\pi_1(j)} , \z_{\pi_2(j)}) \\
\end{align*}
which corresponds exactly to the kernel $\mathcal{K}_w$ defined in Eq. \ref{eq:kernel_def}. This proves that $\mathcal{K}_w$ in Eq. \ref{eq:kernel_def} is a well defined PSD kernel. \\

With the particular choice $w(\mathbf{A}) = \left\langle \mathbf{A},\mathbf{\Omega_{sim}} \right\rangle$, we recover: 
\begin{align*}
\mathcal{K}_w(\y,\z) &=   \sum_{\mathbf{A} \in \mathcal{A}} \left\langle \mathbf{A},\mathbf{\Omega_{sim}} \right\rangle  ~~ e^{ -  \frac{\left\langle \mathbf{A} , \mathbf{\Delta(\y,\z)} \right\rangle}{\gamma} } \\
&= Z  \times \text{TDI}^{\mathbf{\Delta,\Omega_{sim}}}_{\gamma}(\y,\z) \\
&=  e^{- \text{DTW}^{\mathbf{\Delta}}_{\gamma}(\y,\z) / \gamma}
  \times \text{TDI}^{\mathbf{\Delta, {\Omega_{sim}}}}_{\gamma} (\y,\z) \\
  &= \mathcal{K}_{time}(\y,\z)
\end{align*}
which finally proves that $\mathcal{K}_{time}$ defined in paper Eq. 9 is a valid PSD kernel.
\end{proof}{}

The particular choice  $k(u,v)= \dfrac{\frac{1}{2} e^{-\Vert u-v \Vert^2_2}} {1-\frac{1}{2} e^{- \Vert u-v \Vert^2_2}}$ fullfills Proposition 1 requirements: $k$ is indeed PSD as the infinite limit of a sequence of PSD kernels $\sum_{i=1}^{\infty} g^i = \frac{g}{1-g} = k$, where $g$ is a halved Gaussian PSD kernel: $g(u,v)=  \frac{1}{2} e^{- \Vert u-v \Vert ^2_2}$. For this choice of $k$, the corresponding pairwise cost matrix writes (it is the half-Gaussian cost defined in paper section \ref{sec:shape-kernel}):
\begin{equation}
    \mathbf{\Delta}(\y_i,\z_j) = \gamma \left[\Vert \y_i-\z_j\Vert^2_2  - \log \left( 2 - e^{- \Vert \y_i-\z_j \Vert ^2_2}  \right)  \right] 
\end{equation}

\section{Efficient forward and backward computation} 
\label{app:efficient-computation}

The direct computation of the shape loss $\text{DTW}^{\mathbf{\Delta}}_{\gamma}$ (paper Eq. 3) and the temporal loss $\text{TDI}^{\mathbf{\Delta,\Omega_{dissim}}}_{\gamma}$ (paper  Eq. 8) is intractable, due to the exponential growth of cardinal of $\mathcal{A}_{n,m}$. We provide a careful implementation of the forward and backward passes in order to make learning efficient.\\

\textbf{Shape loss:} Regarding $\text{DTW}^{\mathbf{\Delta}}_{\gamma}$, we rely on~\cite{cuturi2017soft} to efficiently compute the forward pass with a variant of the Bellmann dynamic programming approach~\cite{bellman1952theory}. For the backward pass, we implement the recursion proposed in~\cite{cuturi2017soft} in a custom Pytorch loss. This implementation is much more efficient than relying on vanilla auto-differentiation, since it reuses intermediate results from the forward pass.\\

\textbf{Temporal loss:} For $\text{TDI}^{\mathbf{\Delta},\mathbf{\Omega_{dissim}}}_{\gamma}$, note that the bottleneck for the forward pass in Eq. 8 is to compute $\mathbf{A}^*_{\gamma} = \nabla_{\Delta} \text{DTW}^{\mathbf{\Delta}}_{\gamma}(\y,\z)$, which we implement as explained for the  $\text{DTW}^{\mathbf{\Delta}}_{\gamma}$ backward pass. Regarding $\text{TDI}^{\mathbf{\Delta},\mathbf{\Omega_{dissim}}}_{\gamma}$  backward pass, we need to compute the Hessian $\nabla^2 \text{DTW}^{\mathbf{\Delta}}_{\gamma}(\y,\z)$. We use the method proposed in~\cite{mensch2018differentiable}, based on a dynamic programming implementation that we embed in a custom Pytorch loss. Again, our back-prop implementation allows a significant speed-up compared to standard auto-differentiation (see paper section 5.2.3). 
The resulting time complexity of both shape and temporal losses for forward and backward is $\mathcal{O}(nm)$. \\

\begin{figure}
    \centering
    \includegraphics[width=8cm]{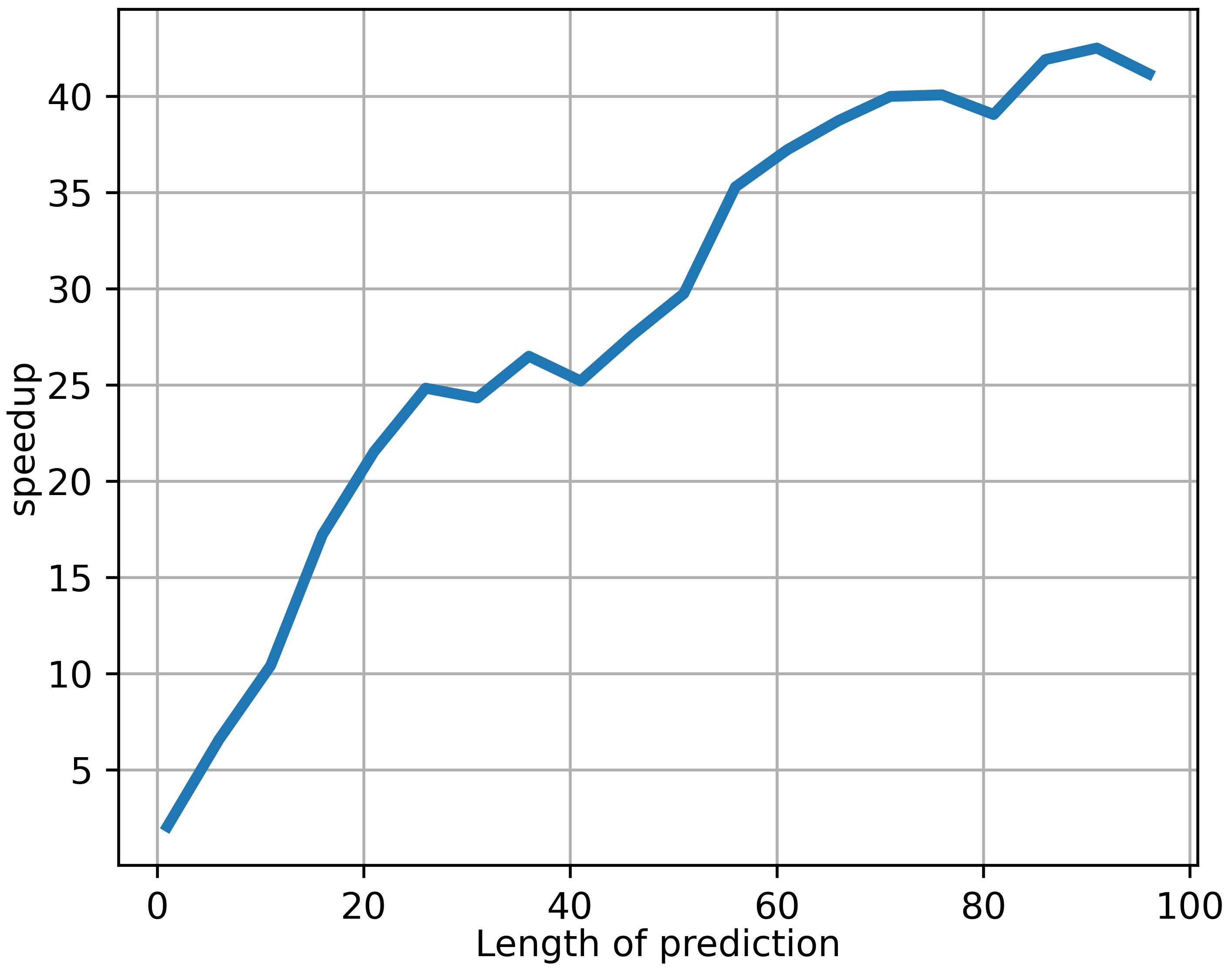}
    \caption{Speedup of the custom forward and backward implementation of DILATE.}
    \label{fig:speedup}
\end{figure}

\textbf{Custom backward implementation speedup}: We compare in Figure \ref{fig:speedup} the computational time between the standard Pytorch auto-differentiation mechanism and our custom backward pass implementation. We plot the speedup of our implementation with respect to the prediction length $\tau$ (averaged over 10 random target/prediction tuples). We notice the increasing speedup with respect to $\tau$: speedup of $\times$ 20 for 20 steps ahead and up to $\times$ 35 for 100 steps ahead predictions.

\section{Datasets}
\label{app:datasets}

We provide here a detailed description of the datasets used in the paper. We precise for each case the length of the context window $T$ and the prediction horizon $\tau$.\\

\textbf{Synthetic-det} ($T=20, \tau=20$): dataset consisting in predicting sudden changes (step functions) based on an input signal composed of two peaks. This controlled setup was designed to measure precisely the shape and time errors of predictions. We generate 500 times series for train, 500 for validation and 500 for test, with 40 time steps: the first 20 are the inputs, the last 20 are the targets to forecast. In each series, the input range is composed of 2 peaks of random temporal position $i_1$ and $i_2$ and random amplitude $j_1$ and $j_2$ between 0 and 1, and the target range is composed of a step of amplitude $j_2-j_1$  and  stochastic position $i_2 + (i_2-i_1)+ randint(-3,3)$. All time series are corrupted by an additive Gaussian white noise of variance 0.01.\\

\textbf{Synthetic-prob} ($T=20, \tau=20$): this is a variant of Synthetic-det where for each input series, we generate 10 different future series of length 20 by adding noise on the step amplitude and localisation. The dataset is composed of $100 \times 10=1000$ time series for each train/valid/test split.\\

\textbf{ECG5000} ($T=84, \tau=56$): this dataset comes from the UCR Time Series Classification Archive \cite{chen2015ucr}, and is composed of 5000 electrocardiograms (ECG) (500 for training, 4500 for testing) of length 140. We take the first 84 time steps (60 \%) as input and predict the last 56 steps (40 \%) of each time series (same setup as in \cite{cuturi2017soft}).\\

\textbf{Traffic} ($T=168, \tau=24$): this dataset is composed of road occupancy rates (between 0 and 1) from the California Department of Transportation (48 months from 2015-2016) measured every 1h. We work on the first univariate series of length 17544 (with the same 60/20/20 train/valid/test split as in \cite{lai2018modeling}), and we train models to predict the 24 future points
given the past 168 points (past week). \\

 \textbf{Electricity} ($T=168, \tau=24$): this dataset consists in hourly electricity consumption measurements (kWh) from 370 customers  (see Figure \ref{fig:dilate_elec_etth1} (a)).\\

 \textbf{ETTh1} \cite{zhou2020informer} ($T=96, \tau=96$):  dataset of hourly Electricity Transformer Temperature measurements, which is an important indicator for electricity grids. This dataset enables to assess the generalization of our approach on much longer term predictions (see Figure \ref{fig:dilate_elec_etth1} (b)).

\section{DILATE additional details}
\label{app:dilate}

\subsection{External shape and temporal metrics}

We detail here the two external metrics used in our experiments to evaluate the shape and temporal errors.\\

\textbf{Ramp score:} The notion of \textit{ramping event} is a major issue for intermittent renewable energy production that needs to be anticipated for electricity grid management. For assessing the performance of trained forecasting models in presence of ramps, the Ramp Score was proposed in \cite{vallance2017towards}. This score is based on a piecewise linear approximation on both input and target time series by the Swinging Door algorithm \cite{bristol1990swinging,florita2013identifying}. The Ramp Score described in \cite{vallance2017towards} is computed as the integral between the unsigned difference of derivatives of both linear approximated series. For assessing only the shape error component, we apply in our experiments the ramp score on the target and prediction series after alignment by the optimal DTW path.\\

\textbf{Hausdorff distance:} Given a set of change points $\mathcal{T}^*$ in the target signal and change points $\hat{\mathcal{T}}$ in the predicted signal, the Hausdorff distance is defined as:
\begin{equation}
\text{Hausdorff}(\mathcal{T}^*,\hat{\mathcal{T}}) :=  \max (  \underset{\hat{t} \in \mathcal{ \hat{T} }}{\max}  \underset{t^* \in \mathcal{ T^* }}{\min} |\hat{t}-t^* |  ,  \underset{t^* \in \mathcal{ T^* }}{\max}   \underset{\hat{t} \in \mathcal{ \hat{T} }}{\min} |\hat{t}-t^* | )
\end{equation}{}

It corresponds to the greatest temporal distance between a change point and its prediction. \\

We now explain how the change points are computed for each dataset: for Synthetic, we know exactly by construction the positions of the change points in the target signals. For the predictions, we look for a single change point corresponding to the location of the predicted step function. We use the exact segmentation method by dynamic programming described in \cite{truong2018review} with the Python toolbox \url{http://ctruong.perso.math.cnrs.fr/ruptures-docs/build/html/index.html#} .\\

For ECG5000 and Traffic datasets which present sharp peaks, this change point detection algorithm is not suited (detected change points are often located at the inflexion points of peaks and not at the exact peak location). We thus use a simple peak detection algorithm based on first order finite differences. We tune the threshold parameter for outputting a detection and the min distance between detections parameter experimentally for each dataset.

\subsection{DILATE implementation details}

\textbf{Neural networks architectures:} For the generic neural network architectures, we use a fully connected network (1 layer of 128 neurons), which does not make any assumption on data structure, and a more specialized Seq2Seq model \cite{sutskever2014sequence} with Gated Recurrent Units (GRU) \cite{cho2014learning} with 1 layer of 128 units. 
Each model is trained with PyTorch for a max number of 1000 epochs with Early Stopping with the ADAM optimizer. The smoothing parameter $\gamma$ of DTW and TDI is set to $10^{-2}$. \\

\textbf{DILATE hyperparameters:} the hyperparameter $\alpha$ balancing $\mathcal{L}_{shape}$ and $\mathcal{L}_{time}$ is determined on a validation set to get comparable DTW shape performance than the $\text{DTW}_{\gamma}^{\mathbf{\Delta}}$ trained model: $\alpha=0.5$  for Synthetic and ECG5000, and 0.8 for Traffic, Electricity and ETTh1. The DTW smoothing parameter $\gamma$ is fixed to $10^{-2}$, as discussed in paper section 5.2.3. \\

Our code implementing DILATE is available on line from \url{https://github.com/vincent-leguen/DILATE}.

\subsection{Visualizations for Electricity and ETTh1}
\label{app:elec-etth1}

\begin{figure*}
    \centering
\rotatebox{90}{\textbf{(a) Electricity}}    
\begin{tabular}{cc}
\textbf{N-Beats \cite{oreshkin2019n} MSE} & \textbf{N-Beats \cite{oreshkin2019n} DILATE} \\
\includegraphics[width=8cm]{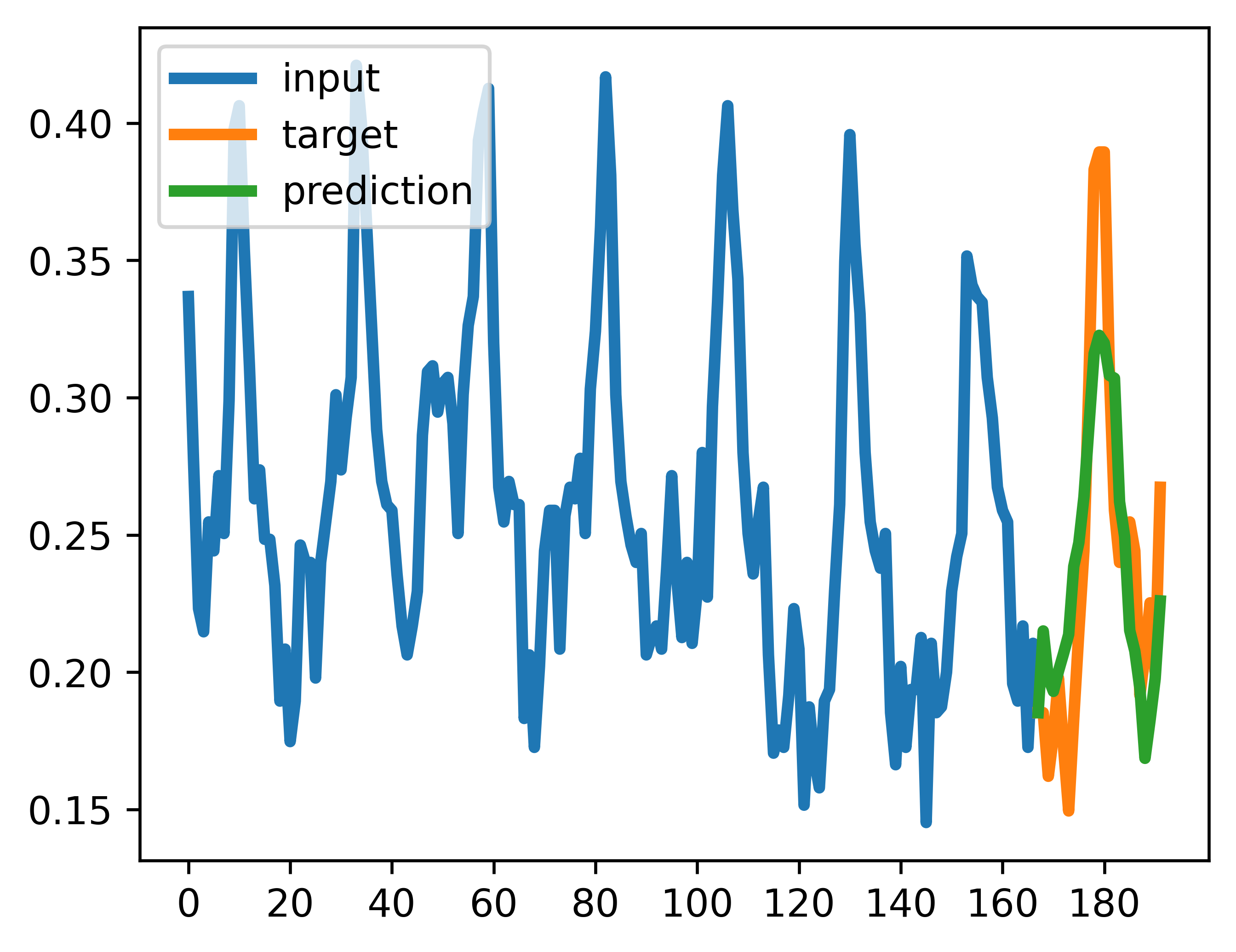}     &  \includegraphics[width=8cm]{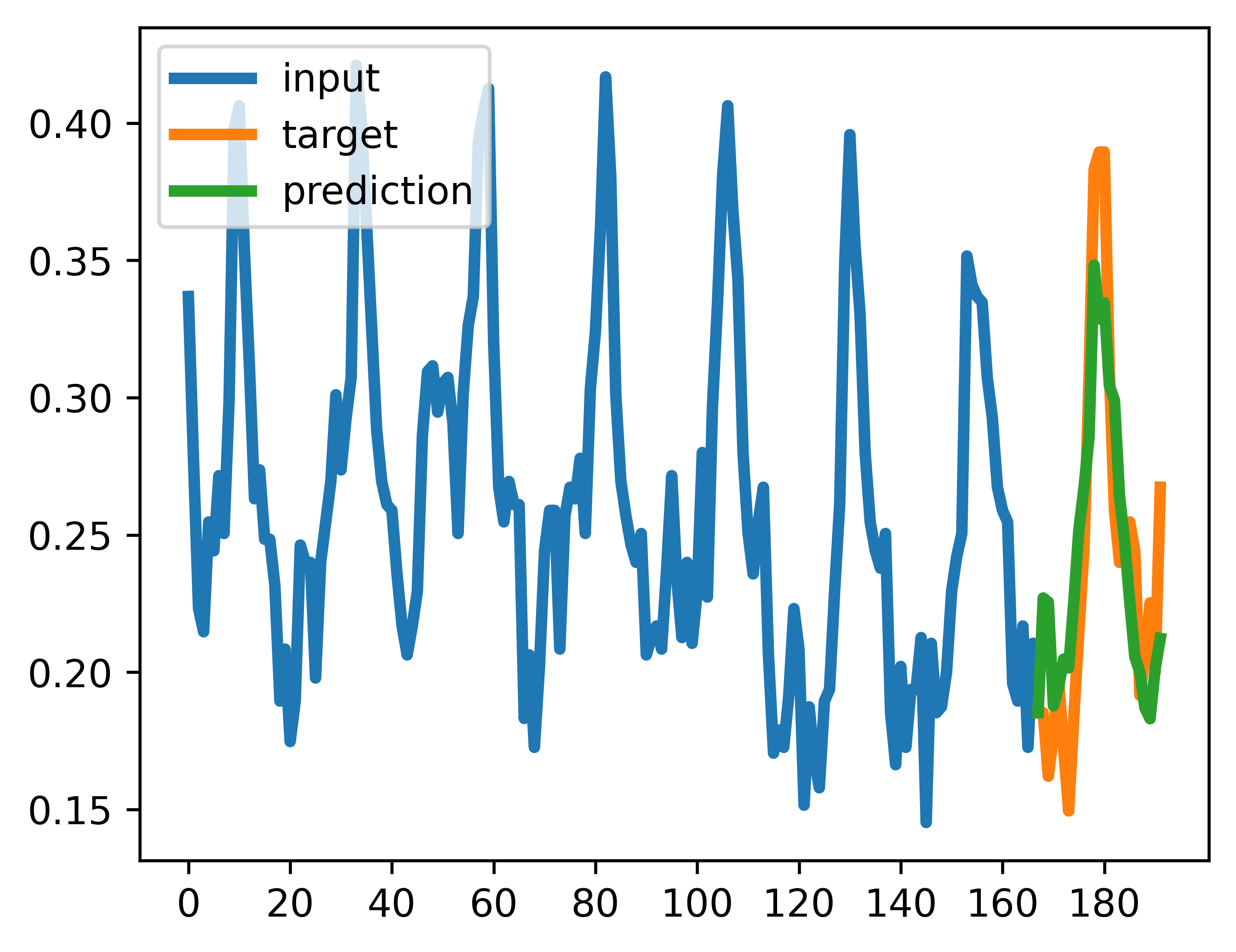} \\
\end{tabular}\\
    \centering
\rotatebox{90}{\textbf{(b) ETTh1}}
\begin{tabular}{cc}
\textbf{N-Beats \cite{oreshkin2019n} MSE} & \textbf{N-Beats \cite{oreshkin2019n} DILATE} \\
\includegraphics[width=8.2cm]{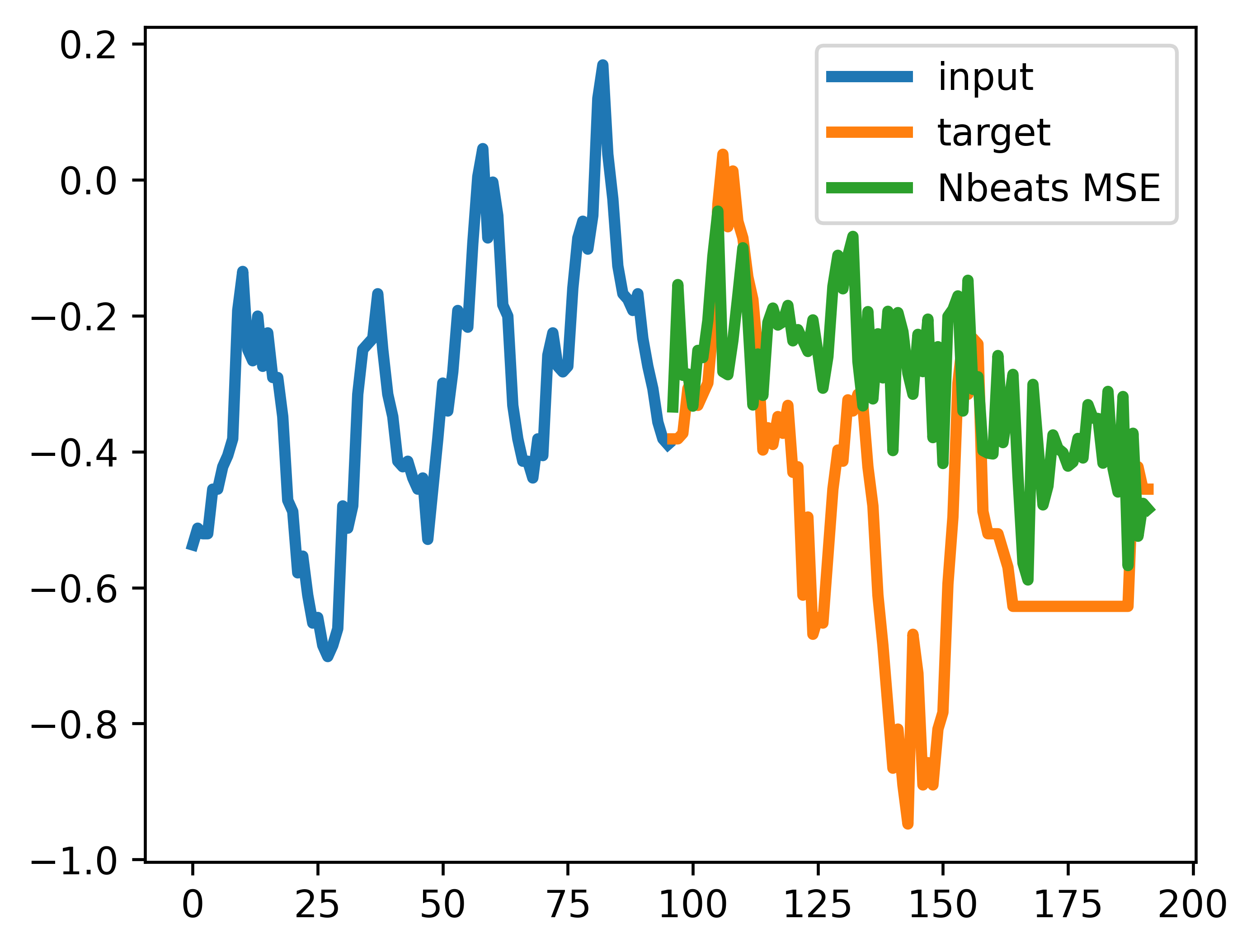}     & 
\includegraphics[width=8.2cm]{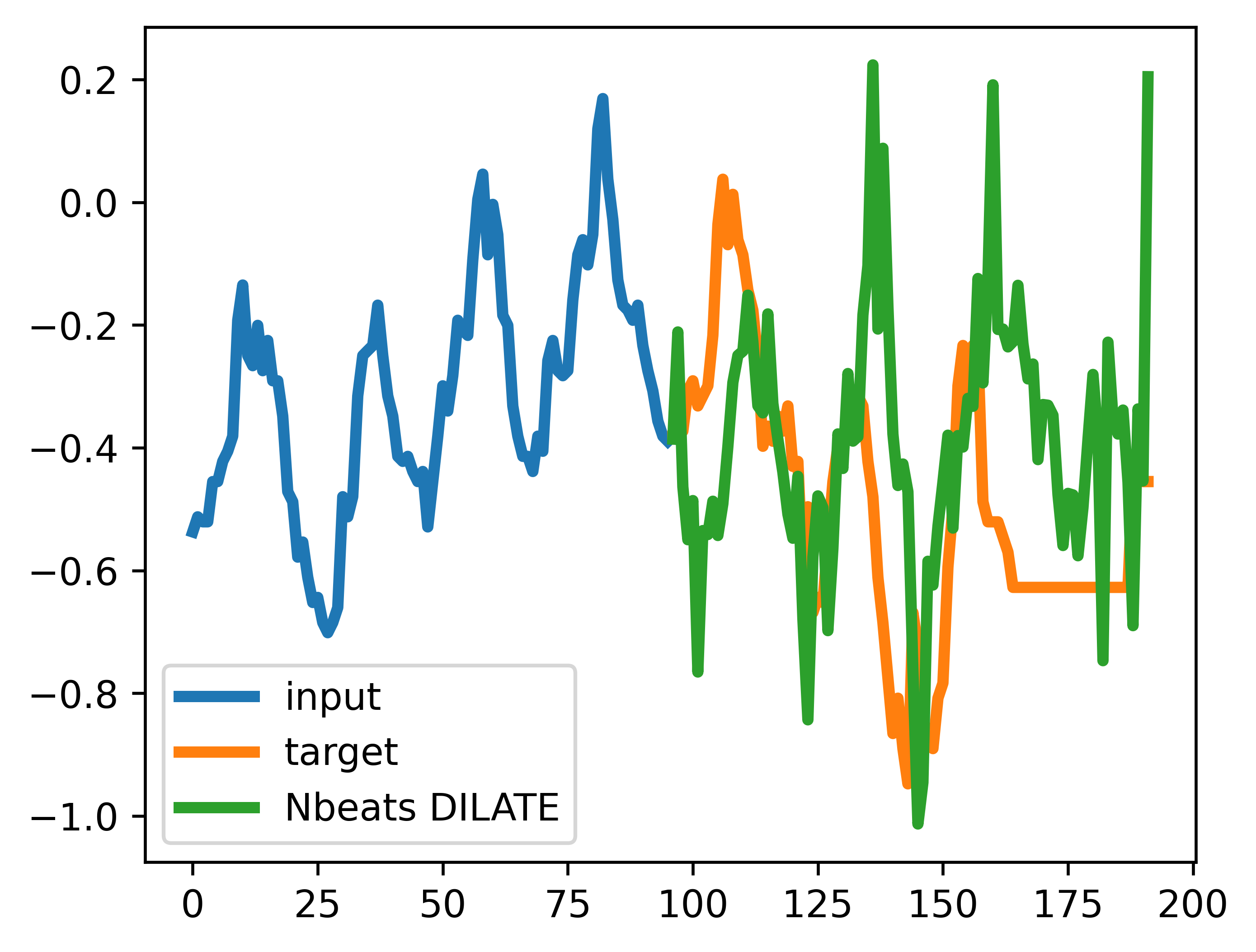}  
\end{tabular}

\caption{Qualitative forecasting results comparing the N-Beats model \cite{oreshkin2019n} trained with MSE and DILATE loss on the Electricity and ETTH1 datasets.}
\label{fig:dilate_elec_etth1}
\end{figure*}

We provide in Figure \ref{fig:dilate_elec_etth1} qualitative predictions of the N-Beats model \cite{oreshkin2019n} on the Electricity (a) and ETTh1 (b) datasets. For Electricity (a), DILATE leads to a much sharper prediction than with the MSE, as expected. For ETTH1 (b), we extend the experiments to a much longer prediction horizon ($\tau=96$), which makes the extrapolation task much more challenging. We observe that the MSE loss only gives a future trend trajectory, while DILATE better captures the future sharp patterns.

\subsection{Comparison to DILATE divergence variant \label{app:dilate-div}}

Blondel \etal \cite{blondel2020differentiable} point out two limitations for using $\text{DTW}^{\mathbf{\Delta}}_{\gamma}$ as a loss function: first, it can take negative values and second, $\text{DTW}^{\mathbf{\Delta}}_{\gamma}(\y,\z)$ does not reach its minimum when $\y = \z$. To address these issues, they propose a proper divergence defined as follows \cite{blondel2020differentiable}:
\begin{multline}
    \text{DTW-div}^{\mathbf{\Delta}}_{\gamma}(\y, \z) =  \text{DTW}^{\mathbf{\Delta}}_{\gamma}(\y, \z) \\ - \frac{1}{2} (\text{DTW}^{\mathbf{\Delta}}_{\gamma}(\y, \y) + \text{DTW}^{\mathbf{\Delta}}_{\gamma}(\z, \z))
\end{multline}
This divergence is non-negative and satisfies $ \text{DTW-div}^{\mathbf{\Delta}}_{\gamma}(\y, \y) = 0$. However, it is still not a distance function since the triangle inequality is not verified (as for the true DTW).\\

These limitations also hold for DILATE. Consequently, we use the same normalization trick to define a proper DILATE-divergence. Forecasting results in Table \ref{tab:dilate-div} show that DILATE-div is equivalent to DILATE with the Seq2Seq and N-Beats \cite{oreshkin2019n} models, and inferior to DILATE with the Informer model \cite{zhou2020informer}. It confirms the good behaviour of the DILATE loss that does not require this renormalization.

\begin{table}[H]
    \caption{Comparison between DILATE and DILATE-div on the synthetic-det dataset.}
    \centering
    \begin{tabular}{ccc}
    \toprule
   Model & MSE  & DILATE    \\
      \midrule
Seq2Seq DILATE &  \textbf{13.1 $\pm$ 1.8} &  \textbf{33.7 $\pm$ 3.1} \\
 Seq2Seq  DILATE-div & \textbf{13.6 $\pm$ 0.9}     & \textbf{33.6 $\pm$ 2.1} \\
   \midrule
  N-Beats \cite{oreshkin2019n} DILATE & \textbf{13.3 $\pm$ 0.7}     & \textbf{37.9 $\pm$ 1.6}  \\
   N-Beats \cite{oreshkin2019n} DILATE-div & \textbf{13.8 $\pm$ 0.9}     & \textbf{38.5 $\pm$ 1.4}  \\
  \midrule
  Informer \cite{zhou2020informer} DILATE & \textbf{11.8 $\pm$ 0.7}    & \textbf{30.1 $\pm$ 1.3} \\
   Informer \cite{zhou2020informer} DILATE-div & 12.9 $\pm$ 0.1   & 31.8 $\pm$ 6.5 \\  
 \bottomrule
    \end{tabular}
    \label{tab:dilate-div}
\end{table}

\subsection{Additional visualizations}
\label{app:dilate_visus}

\begin{figure*}
\begin{center}
 \includegraphics[width=13cm]{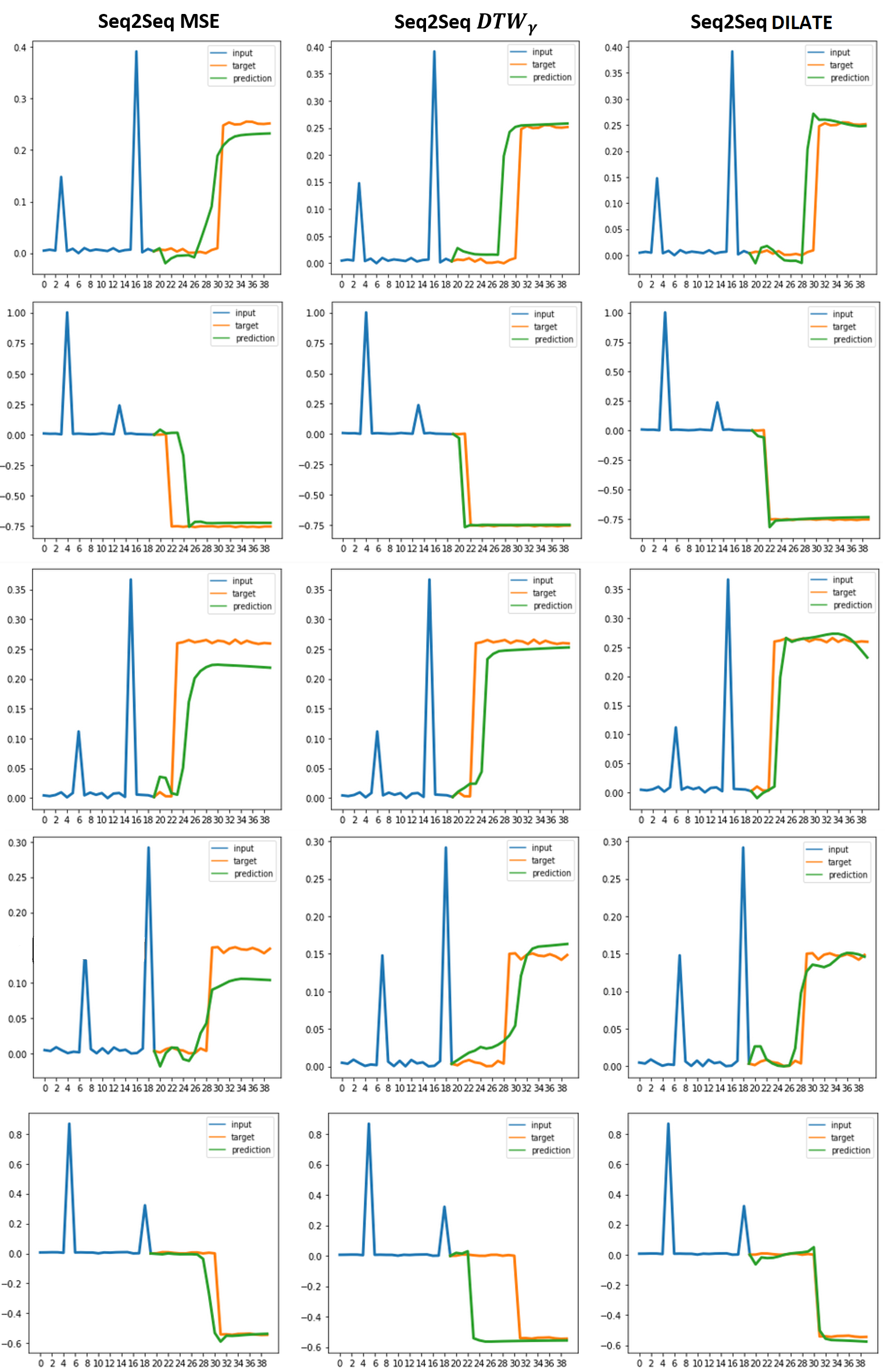}  
\end{center}
   \caption{Qualitative predictions for the Synthetic-det dataset.}
  \label{fig:synth_sup}
\end{figure*}

\begin{figure*}
\begin{center}
 \includegraphics[width=13cm]{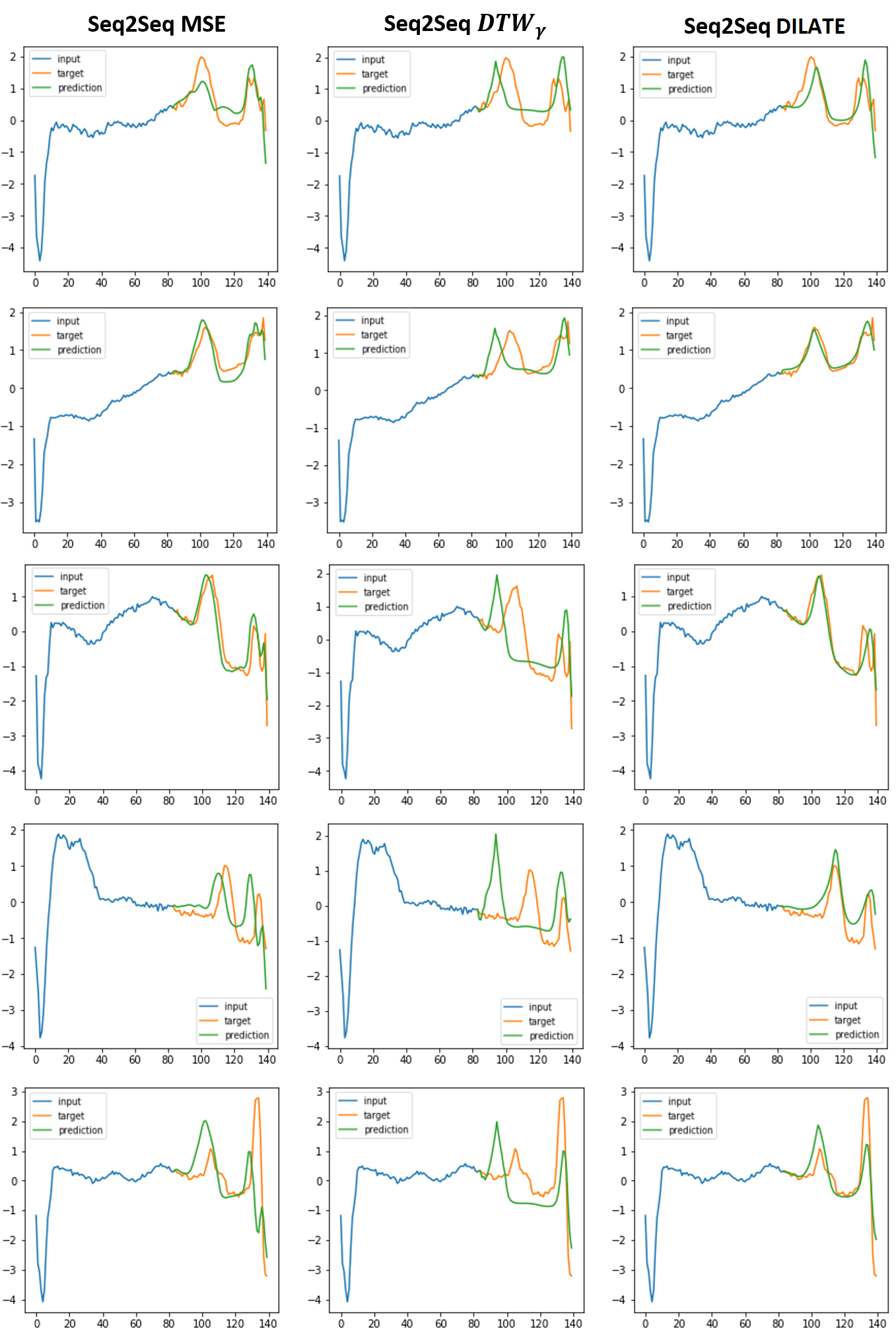}  
\end{center}
   \caption{Qualitative predictions for the ECG5000 dataset.}
  \label{fig:ecg_sup}
\end{figure*}

\begin{figure*}
\begin{center}
 \includegraphics[width=13cm]{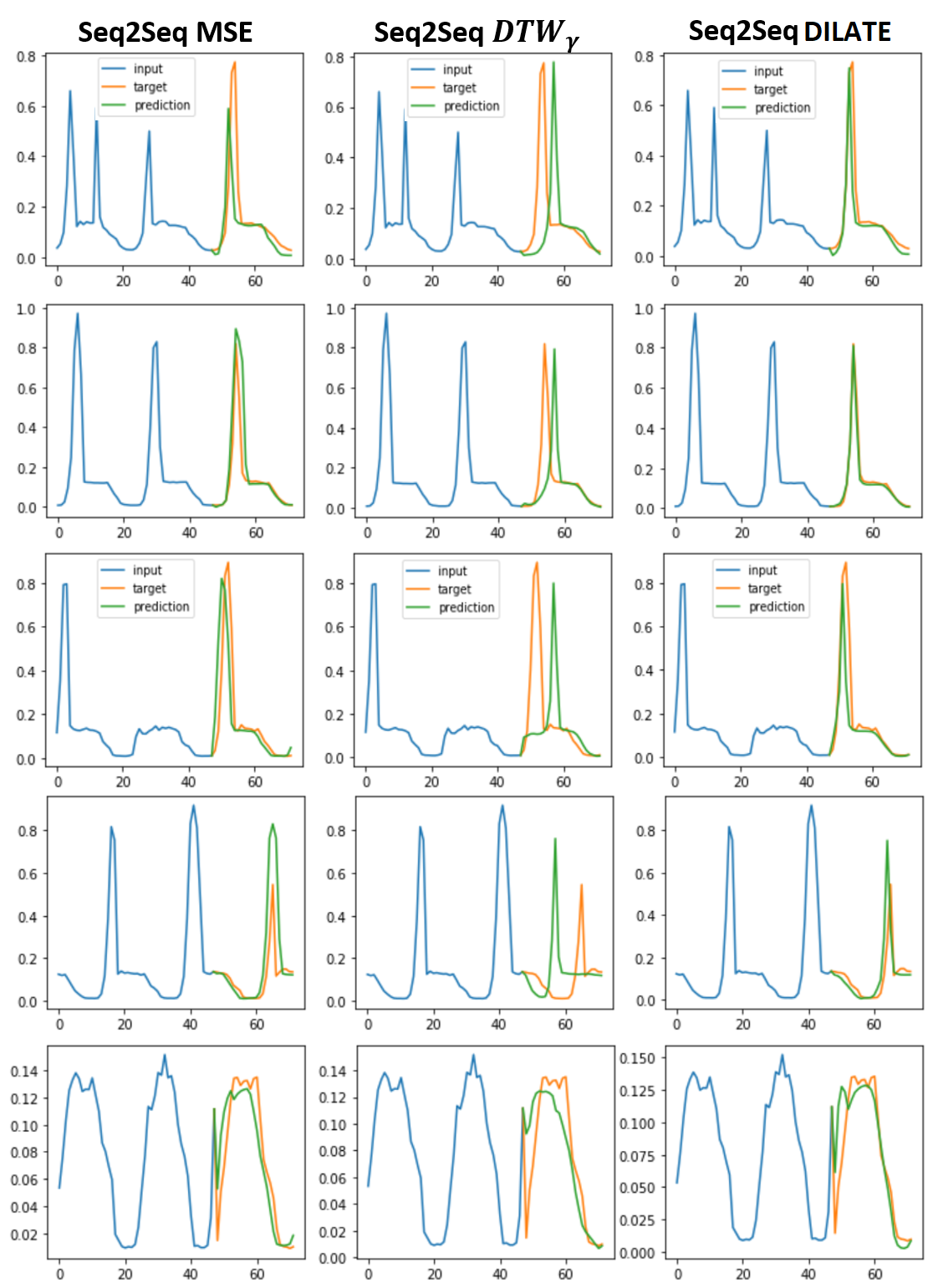}  
\end{center}
   \caption{Qualitative predictions for the Traffic dataset.}
  \label{fig:traffic_sup}
\end{figure*}

We provide additional qualitative predictions with DILATE for the Synthetic-det in Figure \ref{fig:synth_sup}, for ECG5000 in Figure \ref{fig:ecg_sup} and for Traffic in Figure \ref{fig:traffic_sup}.

\pagebreak
\section{STRIPE++ additional details}

\subsection{Derivation of the DPP diversity loss}
\label{app:dpp}

Determinantal Point Processes (DPPs) \cite{kulesza2012determinantal} are a probabilistic tool for describing the diversity of a ground set of items $\mathcal{S}= \left\{\mathbf{y}_1,...,\mathbf{y}_N \right\}$. Diversity is controlled via the choice of a positive semi-definite (PSD) kernel $\mathcal{K}$ for comparing items. A DPP is a probability distribution over all subsets of $\mathcal{S}$ that assigns the following probability to a random subset $\mathbf{Y}$:
\begin{equation}
    \mathcal{P}_{\mathbf{K}}(\mathbf{Y}=Y) = \frac{\det(\mathbf{K}_Y)}{\sum_{Y' \subseteq \mathcal{S}}\det(\mathbf{K}_Y')} = \frac{\det(\mathbf{K}_Y)}{\det(\mathbf{K+I})}
\end{equation}{}
where $\mathbf{K}$ denotes the kernel in matrix form and $\mathbf{K}_A$ is its restriction to the elements indexed by $A$ : $\mathbf{K}_A = [\mathbf{K}_{i,j}]_{i,j \in A}$. 

Intuitively, a DPP encourages the selection of diverse elements from the ground set $\mathcal{Y}$. If $\mathcal{Y}$ is more diverse, a random subset $Y \sim \text{DPP}(\mathcal{K})$ sampled from the DPP will select more items, \ie will have a larger cardinality. This idea is embedded into the diversity loss $\mathcal{L}_{diversity}$ proposed in \cite{yuan2019diverse}:
\begin{equation}
    \mathcal{L}_{diversity}(\mathcal{Y} ; \mathbf{K}) = -\mathbb{E}_{Y \sim \text{DPP}(\mathbf{K})} |Y| = - Trace(\mathbf{I}-(\mathbf{K}+\mathbf{I})^{-1})
    \label{eq:ldiversity}
\end{equation}{}

\subsection{STRIPE++ implementation details}
\label{app:stripe-implem}

\textbf{Neural network architectures:} STRIPE++ is composed of a Sequence To Sequence predictive model. The encoder is a recurrent neural network (RNN) with 1 layer of 128 Gated Recurrent Units (GRU) \cite{cho2014learning} units, producing a latent state $h$ of dimension 128. We fixed by cross-validation the dimension of each diversifying variable $z_s$ or $z_t$ to be $k=8$. The decoder is another RNN with $128+8+8=144$ GRU units followed by fully connected layers responsible for producing the future trajectory.\\

The Posterior network has a similar architecture as the encoder: it is a RNN with 1 layer of 128 GRU units that takes as input the full series $(\x_{1:T},\y^*_{T+1:T+\tau})$, followed by two multi-layer perceptrons (MLP) dedicated to output the parameters $(\mu_s^*,\sigma_s^*)$ and $(\mu_t^*,\sigma_t^*)$ of the Gaussian distribution from which to sample the posterior diversifying variables $z_s^*$ and $z_t^*$.\\

The STRIPE$^{++}_{\text{shape}}$ and STRIPE$^{++}_{\text{time}}$ proposal mechanisms build on top of the encoder (that produces $h$) with a MLP with 3 layers of 512 neurons (with Batch Normalization and LeakyReLU activations) and a final linear layer to produce $N=10$ latent codes of dimension $k=8$ (corresponding to the proposals for $z_s$ or $z_t$).\\

\textbf{STRIPE++ hyperparameters:} We cross-validated the relevant hyperparameters of STRIPE++:
\begin{itemize}
\setlength{\itemsep}{5pt}
  \setlength{\parskip}{0pt}
  \setlength{\parsep}{0pt}

    \item $k$: dimension of the diversifying latent variables $z$. This dimension should be chosen relatively to the hidden size of the RNN encoders and decoders (128 in our experiments). We fixed $k=8$ in all cases.
    \item $N$: the number of future trajectories to sample. We fixed $N=10$. We performed a sensibility analysis to this parameter in paper Figure 8.
    \item $\mu = 20$: quality constraint hyperparameter in the DPP kernels.
\end{itemize}

\subsection{STRIPE++ additional visualizations}

\begin{figure}[H]
\centering
\begin{tabular}{c}
   \includegraphics[width=8.5cm]{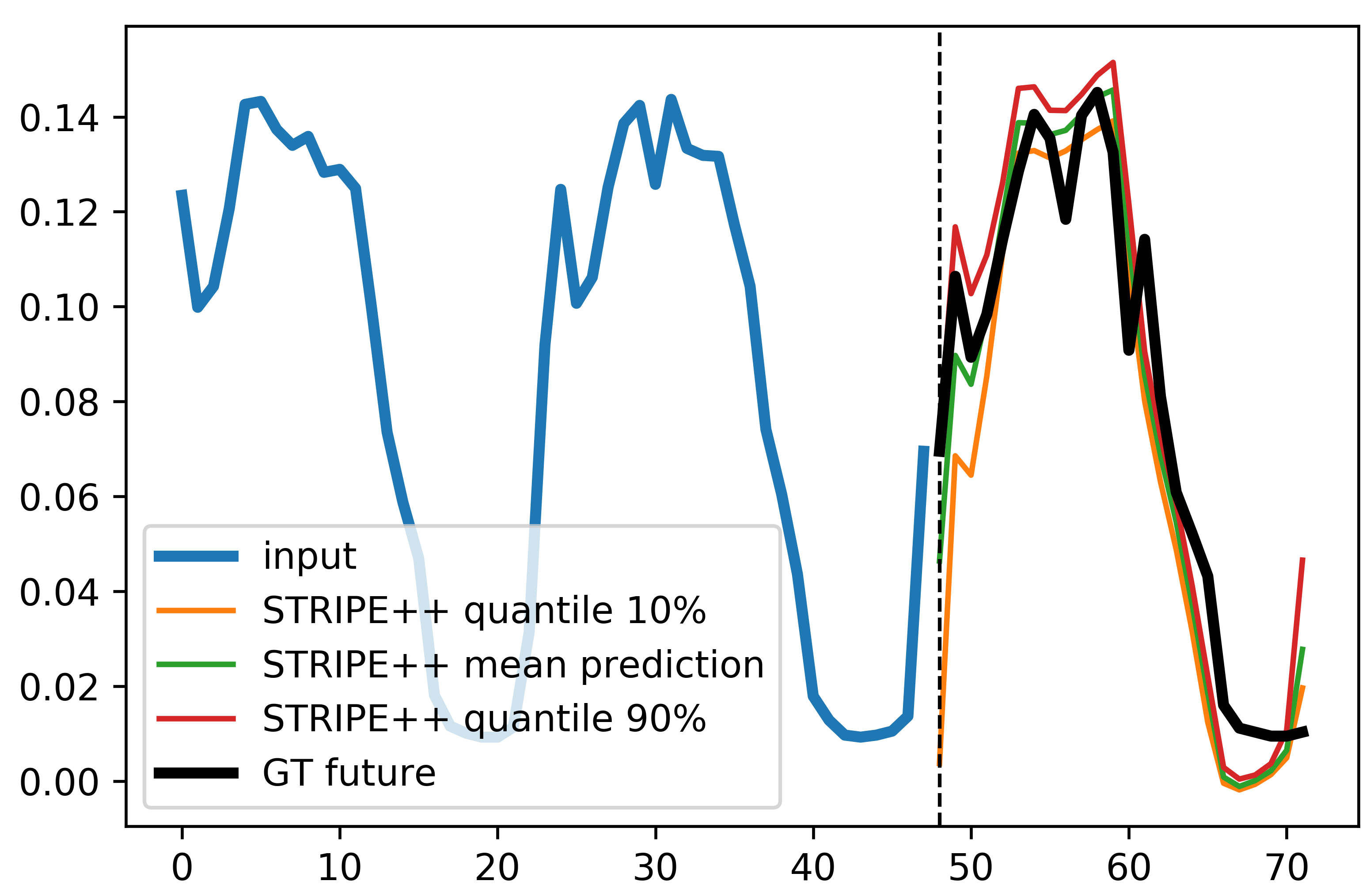}   \\  
   (a) \texttt{Traffic} \\
   \includegraphics[width=8.5cm]{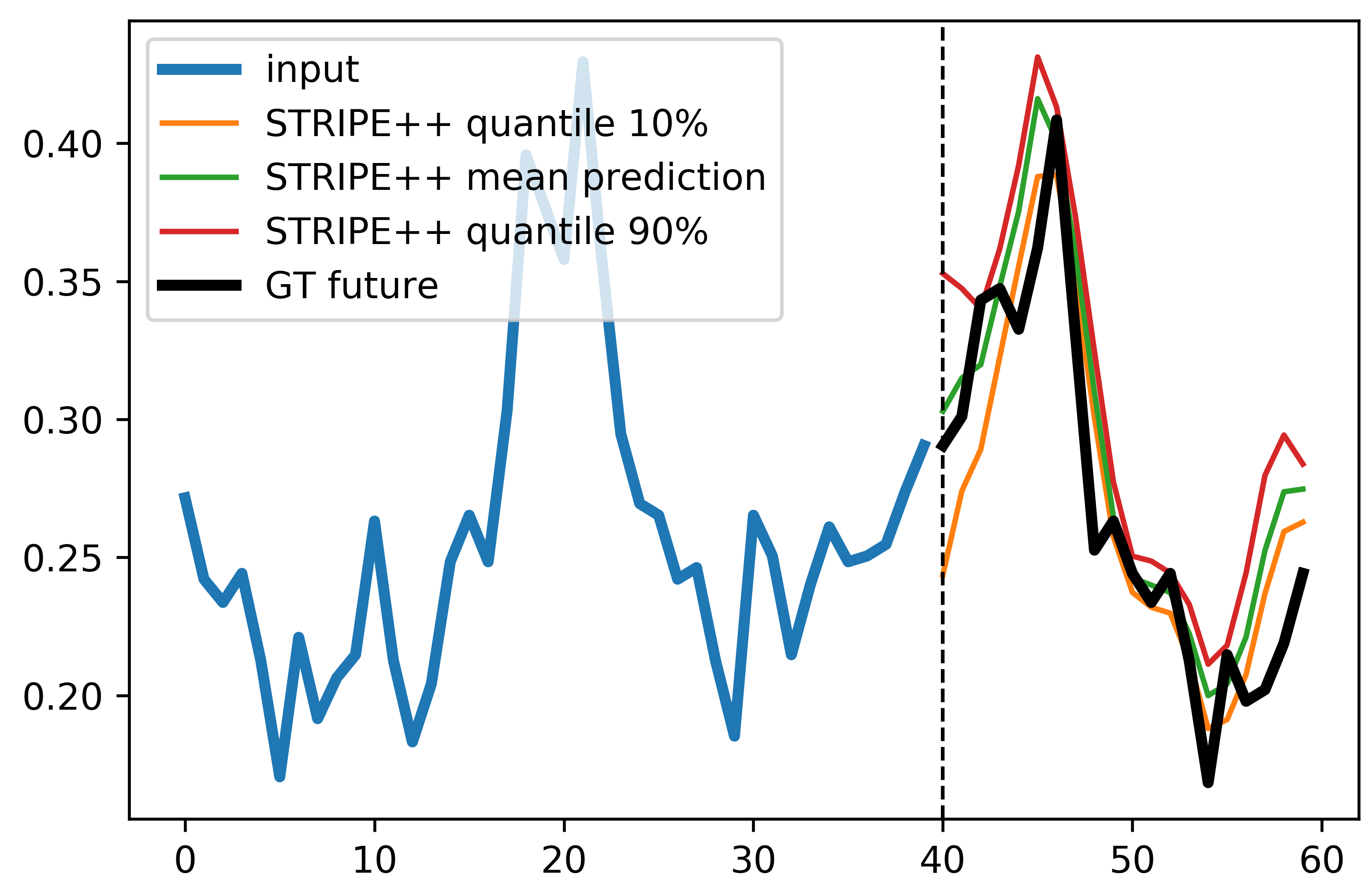}\\
   (b) \texttt{Electricity}
\end{tabular}
    \caption{STRIPE++ qualitative predictions on real-world datasets \texttt{Traffic} (a) and \texttt{Electricity} (b).}
    \label{fig:supp_stripe_visus}
\end{figure}

In paper Figure 7, we have shown qualitative prediction results of the STRIPE++ model on the \texttt{Traffic} and \texttt{Electricity} datasets. To complement these visualizations, we represent in Figure \ref{fig:supp_stripe_visus} for the same examples the 10$^{th}$, 90$^{th}$ quantiles and the mean STRIPE++ prediction, computed over a set of 100 predictions. We observe that the  10$^{th}$ and 90$^{th}$ of STRIPE++ predictions offer a realistic and sharp cover of possible trajectories, that includes the ground truth future.

This contrasts with state-of-the-art deep probabilistic forecasting methods trained with the MSE or the quantile loss \cite{yuan2019diverse,wen2017multi}; the predictions of these models are often smooth and non-sharp, as illustrated in paper Figure \ref{fig-intro} (b). Therefore they do not correspond to realistic scenarios.

\subsection{STRIPE++ quality/diversity cooperation analysis}
\label{app:diverse_dpp}

\begin{figure}
    \centering
    \includegraphics[width=8cm]{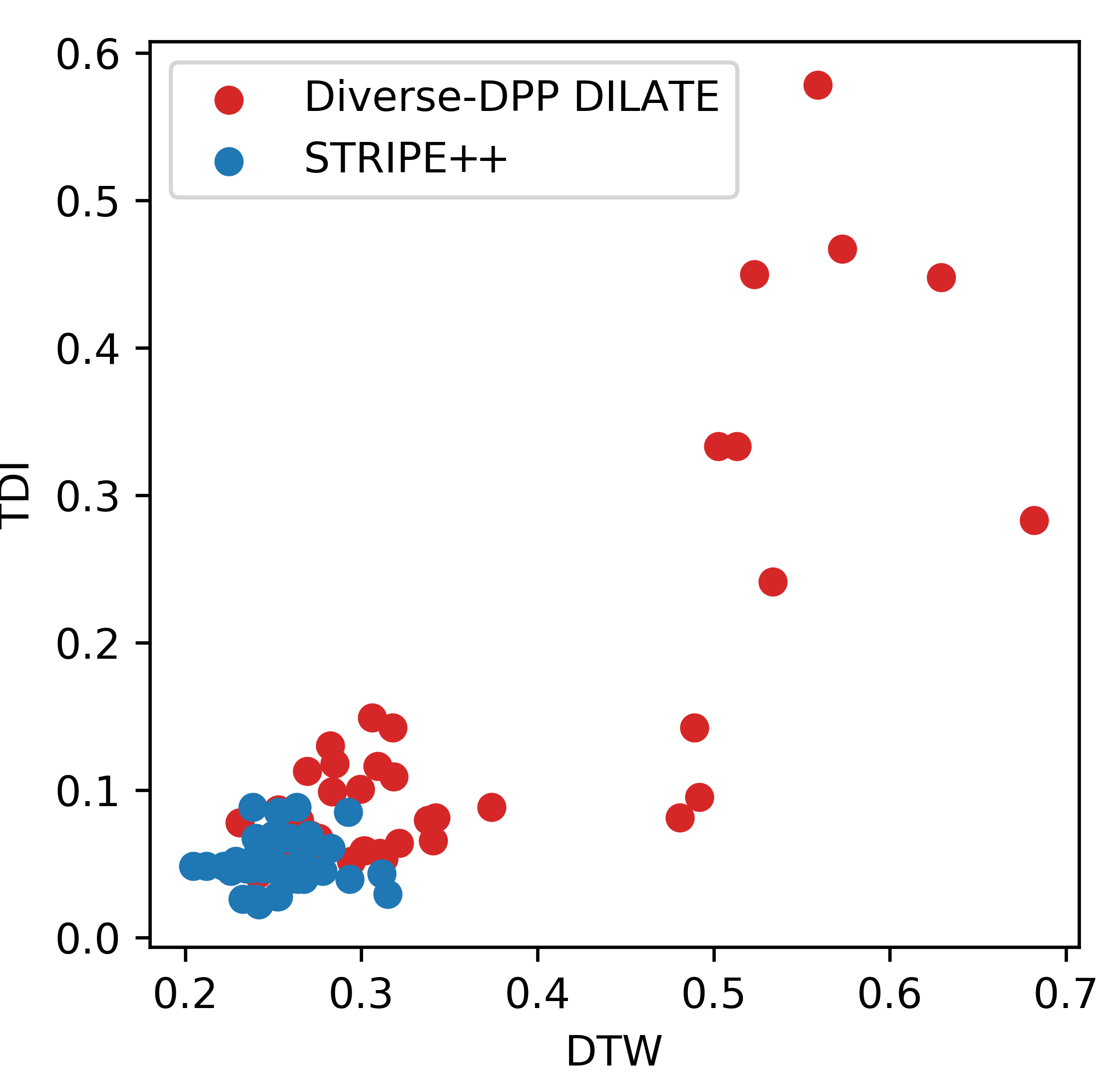}
\caption{Scatterplot of 50 predictions in the plane (DTW,TDI), comparing STRIPE++ v.s. Diverse DPP DILATE \cite{yuan2019diverse}.}
    \label{fig:stripe_scatter}
\end{figure}

We highlight here the importance to separate the criteria for enforcing quality and diversity. In Figure \ref{fig:stripe_scatter}, we represent 50 predictions from the models Diverse DPP DILATE \cite{yuan2019diverse} and STRIPE++ in the plane (DTW,TDI). Diverse DPP DILATE \cite{yuan2019diverse} uses a DPP diversity loss based on the DILATE kernel, which is the same than for quality. We clearly see that the two objectives conflict: this model increases the DILATE diversity (by increasing the variance in the shape (DTW) or the time TDI) components) but a lot of these predictions have a high DILATE loss (worse quality). In contrast, STRIPE++ predictions are diverse in DTW and TDI, and maintain an overall low DILATE loss. STRIPE++ succeeds in recovering a set of good tradeoffs between shape and time leading a low DILATE loss. 

\begin{figure}
    \centering
    \includegraphics[width=9cm]{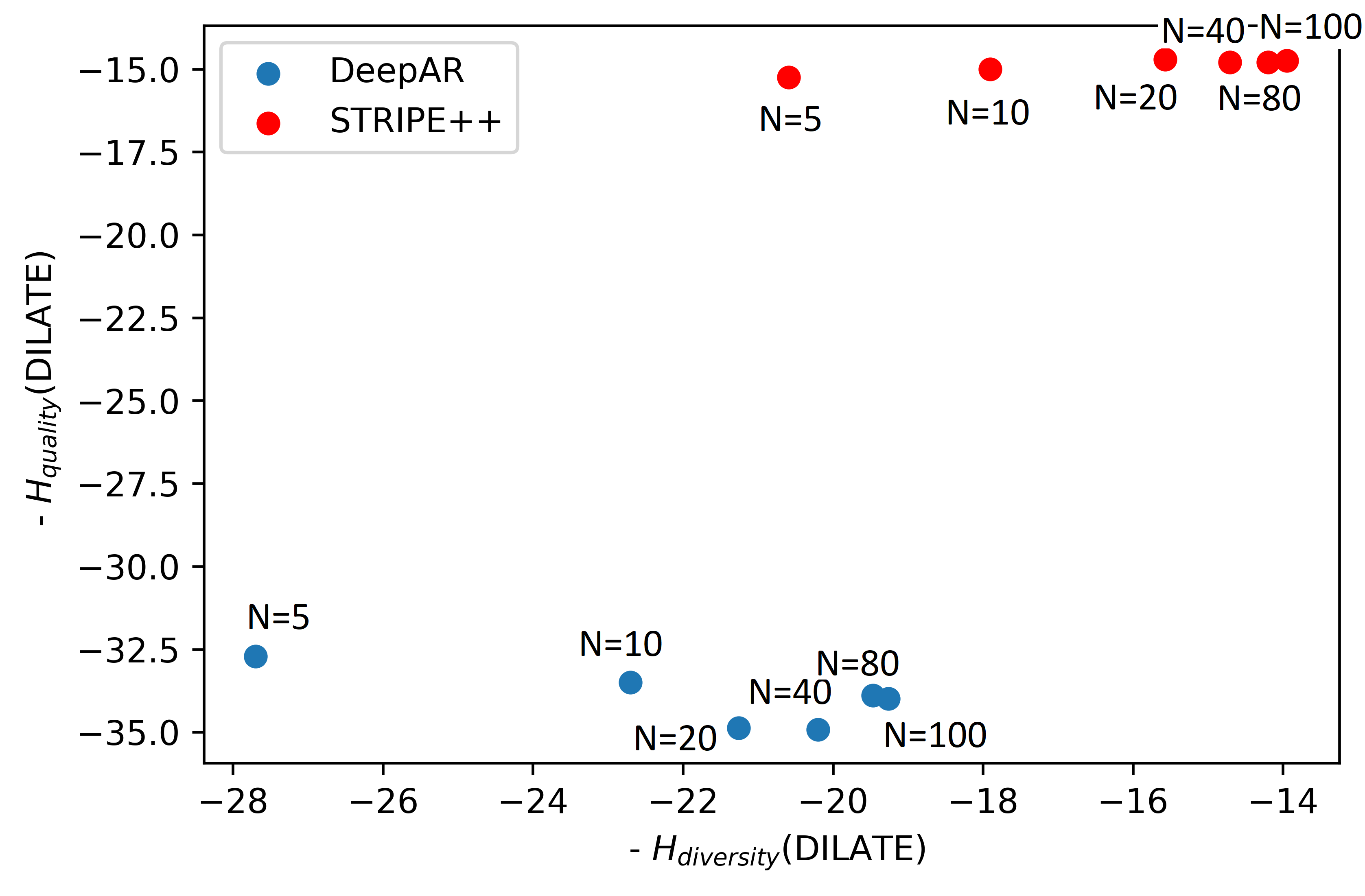}
\caption{Influence of the number $N$ of trajectories on quality ($-\text{H}_{quality}(\text{DILATE)}$, higher is better) and diversity ($-\text{H}_{diversity}(\text{DILATE)}$, higher is better) for the \texttt{synthetic-prob} dataset.}    
    \label{fig:influ_N}
\end{figure}

We also display in Figure \ref{fig:influ_N} the unnormalized results of the comparison between DeepAR \cite{salinas2017deepar} and STRIPE++ with the number of sampled trajectories $N$. The quality is measured with $-\text{H}_{quality}(\text{DILATE)}$ and the diversity with $-\text{H}_{diversity}(\text{DILATE)}$ (higher is better in both cases). These results confirm the conclusions from paper Figure \ref{fig:stripe_N}. The diversity increases for deepAR and STRIPE++ with $N$; however, this gain in diversity comes at the cost of a loss of quality for deepAR, in contrast to STRIPE++ which does not deteriorate quality.

\ifCLASSOPTIONcaptionsoff
  \newpage
\fi




\bibliographystyle{plain}
\bibliography{refs.bib}
